\documentclass{article} 
\usepackage{iclr2024_conference,times}

\usepackage{amsmath,amsfonts,bm}

\def\eqref#1{equation~\ref{#1}}

\def\1{\bm{1}}

\DeclareMathAlphabet{\mathsfit}{\encodingdefault}{\sfdefault}{m}{sl}
\SetMathAlphabet{\mathsfit}{bold}{\encodingdefault}{\sfdefault}{bx}{n}

\usepackage{hyperref}
\usepackage{url}
\usepackage{microtype}
\usepackage{booktabs}
\usepackage{graphicx}
\usepackage{multirow}
\usepackage{amsmath}
\usepackage{xcolor}
\usepackage{tcolorbox}
\usepackage{CJKutf8}

\tcbuselibrary{breakable}
\tcbset{
  breakable,
  break at={\textheight-2\baselineskip},
  break automata/.code={\bottomrule},
  before={\par\noindent\nobreak}
}

\newtcolorbox{mybox}{
  colback=red!10, 
  colframe=red!75!black, 
  boxrule=1pt, 
  arc=4pt, 
  boxsep=0pt, 
  left=6pt, 
  right=6pt, 
  top=6pt, 
  bottom=6pt, 
}

\newtcolorbox{mybox1}{colback=gray!20}

\usepackage{xcolor}
\usepackage{soul}

\definecolor{myblue}{rgb}{0.651, 0.808, 0.890}
\definecolor{mygreen}{rgb}{0.698, 0.875, 0.541}
\definecolor{myorange}{rgb}{0.992, 0.749, 0.435}

\author{%
    Yauwai Yim$^1$\thanks{The first two authors contributed equally.} \hskip1em Chunkit Chan$^1$\footnotemark[1]  
    \hskip1em Tianyu Shi$^2$ \hskip1em \textbf{Zheye Deng$^1$} \hskip1em \\ 
    \textbf{Wei Fan$^1$} \hskip1em \textbf{Tianshi Zheng$^1$} \hskip1em \textbf{Yangqiu Song$^1$} \\ \\
        $^1$The Hong Kong University of Science and Technology\\
        $^2$University of Toronto\\
        \texttt{ywyimaa@connect.ust.hk}, \quad \texttt{yqsong@cse.ust.hk}
}

\iclrfinalcopy 
\begin{document}

\title{Evaluating and Enhancing LLMs Agent based on Theory of Mind in Guandan \begin{CJK}{UTF8}{gkai}(掼蛋)\end{CJK}: A Multi-Player Cooperative Game under Imperfect Information}

\maketitle

\begin{abstract}
\begin{CJK}{UTF8}{gkai}
Large language models (LLMs) have shown success in handling simple games with imperfect information and enabling multi-agent coordination, but their ability to facilitate practical collaboration against other agents in complex, imperfect information environments, especially in a non-English environment, still needs to be explored. This study investigates the applicability of knowledge acquired by open-source and API-based LLMs to sophisticated text-based games requiring agent collaboration under imperfect information, comparing their performance to established baselines using other types of agents. We propose a Theory of Mind (ToM) planning technique that allows LLM agents to adapt their strategy against various adversaries using only game rules, current state, and historical context as input. An external tool was incorporated to mitigate the challenge of dynamic and extensive action spaces in this card game. Our results show that although a performance gap exists between current LLMs and state-of-the-art reinforcement learning (RL) models, LLMs demonstrate ToM capabilities in this game setting. It consistently improves their performance against opposing agents, suggesting their ability to understand the actions of allies and adversaries and establish collaboration with allies. To encourage further research and understanding, we have made our codebase openly accessible.
\end{CJK}
\end{abstract}

\section{Introduction}
Recent progress in Large Language Models (LLMs)~\citep{ DBLP:journals/corr/abs-2307-09288, DBLP:journals/jmlr/ChowdheryNDBMRBCSGSSTMRBTSPRDHPBAI23, DBLP:conf/eacl/ChanCWJFLS24}, has demonstrated their potential in solving complex problems and the LLM-based agent harnesses the capability to complete tasks in some complicated environments \citep{DBLP:books/ox/22/WangWEPTK22, DBLP:journals/corr/abs-2211-09935, DBLP:journals/corr/abs-2305-15486}. However, these methods have several significant limitations: \textbf{(1)} The assumption that the agent has access to all necessary information, which is often different in real-world situations. Games such as diplomacy~\citep{Bakhtin2022HumanlevelPI, DBLP:conf/iclr/GrayLBB21} and poker~\citep{Moravck2017DeepStackEA, Brown2018SuperhumanAF} underscore the difficulties posed by imperfect information. \textbf{(2)} Do not require cooperation among agents, which is uncommon in most scenarios, such as Overcooked~\citep{DBLP:conf/nips/CarrollSHGSAD19}, agents must collaborate to complete their objectives. \textbf{(3)} All tests in an English language environment.

Despite these limitations, some researchers have made progress in addressing them. For instance, ~\citep{li2023theory, DBLP:journals/corr/abs-2310-03903} have demonstrated the ability of LLM-based agents to cooperate and complete tasks in perfect information settings, while~\cite{Suspicion-Agent} has shown that LLM agents can perform well in imperfect information games but without requiring cooperation. However, the capabilities of LLM agents in imperfect information games and scenarios necessitating cooperation are still under-explored. Furthermore, the research on LLM-based agents in non-English environments presents a pressing demand to catch up to that in English-language settings. 

To evaluate the performance of LLMs in imperfect information environments that require collaboration, we selected Guandan\footnote{https://en.wikipedia.org/wiki/Guandan}, a popular Chinese card game. The game's objective is for two players on the same team to cooperate and play out all the cards in their hands as quickly as possible to achieve a higher team score (see Appendix~\ref{app:a} for detailed rules). In alignment with recent work, we evaluated commercial and Chinese open-source models without fine-tuning in a zero-shot setting against various agents, including the state-of-the-art RL agent Danzero+~\citep{zhao2023danzero+}, to assess their adaptability and performance in an imperfect information environment requiring collaboration.

Our evaluation found that LLMs performed poorly in this environment compared to Danzero+. To address this gap, we first analyzed the models' performance and discovered a significant need for LLM in handling variable and lengthy valid action lists. In Guandan, the valid actions for each agent's turn are not fixed, sometimes exceeding 80 actions when a player has the right to play. LLMs struggle to thoroughly analyze all action elements, especially the elements near the end of lists. To mitigate this issue, we introduced an externally trained RL-based tool that LLM agents can invoke to reduce the number of valid actions that need to be analyzed. 

Inspired by LLMs having exhibited advanced reasoning capabilities and indications of Theory of Mind~\citep{DBLP:journals/corr/abs-2302-02083}, being particularly adept at understanding others' intent and making informed decisions in open-ended scenarios when given specific prompts. We then design a planning method that enables LLMs to develop a Theory of Mind (ToM) for playing Guandan. This approach empowers LLMs by understanding others' beliefs and behavioral patterns without specialized training or examples, as illustrated in Figure~\ref{figure1}. By employing a first-order ToM planning method, LLMs can adjust their strategies based on their understanding of teammates' and opponents' beliefs and behavioral patterns. The results demonstrate that the planning framework enhances the game performance of nearly all models. With the assistance of the action recommender and ToM ability, the GPT-4 model achieves a level of play comparable to the state-of-the-art RL model. Our contributions are as follows: 

 \begin{figure}[t]
  \vspace{-1.3cm}
  \centering
  \includegraphics[width=1\textwidth]{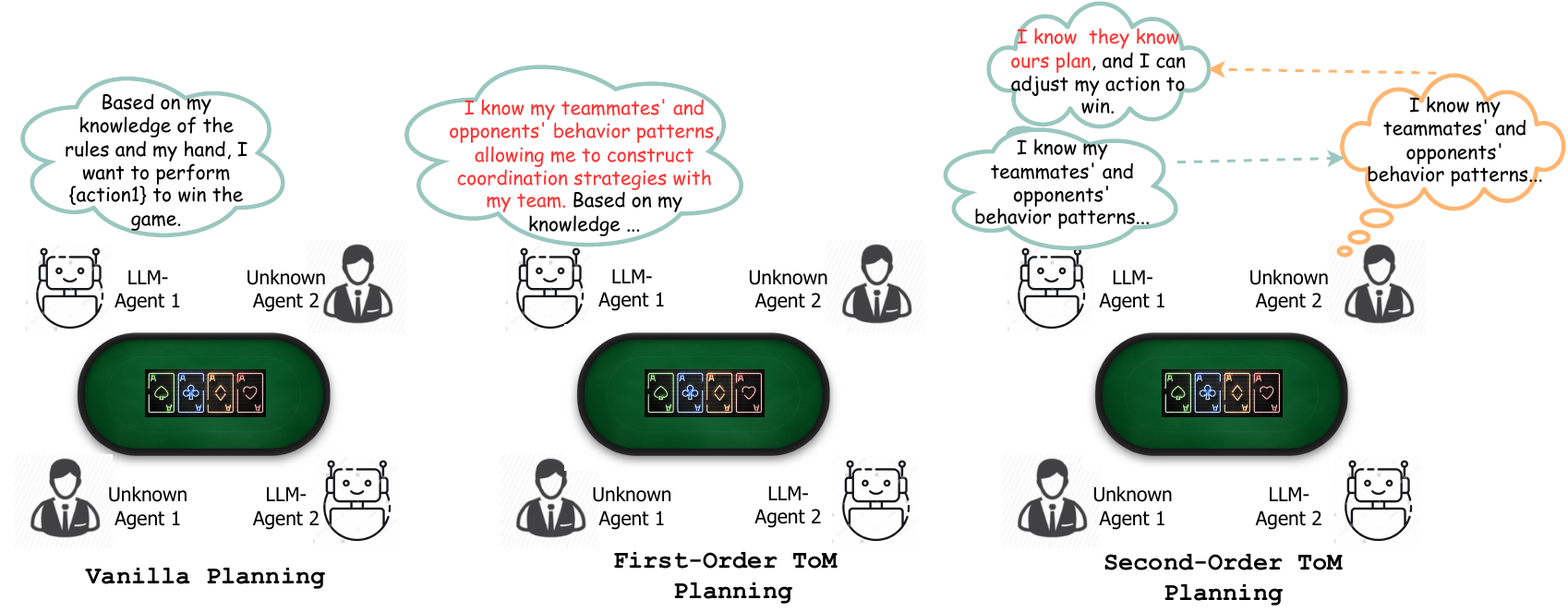}
  \vspace{-0.5 cm}
  \caption{\textbf{Left Figure:} LLM-Agents' reasoning, using vanilla planning.
\textbf{Middle Figure:} LLM-Agents' reasoning, using first-order Theory of Mind(ToM) planning.
\textbf{Right Figure:} LLM-Agents' reasoning, using second-order Theory of Mind(ToM) planning.}
  \label{figure1}
  \vspace{-0.3cm}
\end{figure}

\begin{itemize}

  \item We first conducted a detailed assessment of the performance of both open-source and closed-source LLM agents in a realistic and complex card game, Guandan, which is a Chinese environment that involves imperfect information and necessitates cooperation among agents.
  
  \item We developed and trained an RL-based model that functions as an external tool, which can be seamlessly integrated into the decision-making process of LLM agents, effectively mitigating the challenge of dealing with dynamic and extensive action spaces in those card games.
  
  \item We developed an agent framework that endows these models with different order of Theory of Mind (ToM) capabilities by experimenting with various Chinese and English prompt templates, enabling them to cooperate effectively in this imperfect information game without requiring fine-tuning. Our empirical findings demonstrate that nearly all models exhibited consistent performance improvements after incorporating this type of knowledge.

\end{itemize}

\section{Related work}

\subsection{LLM Reasoning and Multi-agent Collaboration in Imperfect Information Games}

Recent advancements in Large Language Models (LLMs) have showcased their impressive reasoning, planning, and problem-solving capabilities across various tasks~\citep{vicuna2023, GPT4-report, DBLP:conf/nips/Ouyang0JAWMZASR22, DBLP:conf/nips/Wei0SBIXCLZ22, DBLP:conf/nips/YaoYZS00N23}. LLMs augmented with memory, belief, and tools have excelled in multi-step reasoning and outperformed state-of-the-art reinforcement learning methods in open-domain survival games and simple two-player imperfect information games~\citep{DBLP:conf/uist/ParkOCMLB23, DBLP:conf/icml/HuangAPM22, Suspicion-Agent}. Researchers have also explored multi-agent coordination using unreal benchmark environments~\citep{DBLP:conf/nips/CarrollSHGSAD19, DBLP:books/ox/22/WangWEPTK22, DBLP:journals/corr/abs-2310-03903, li2023theory} and studied human collaborative behaviors~\citep{Riedl2021QuantifyingCI}. However, using LLMs to investigate collaborative behaviors in imperfect information games, especially in the Chinese environment, remains largely unexplored. Our research focuses on integrating theory-of-mind (ToM) capabilities into the planning process in Guandan, leveraging LLMs' zero-shot capabilities and ability to facilitate natural language interactions, which is lacking in other studies.

\subsection{Theory of Mind}

Theory of Mind (ToM), a crucial cognitive ability that enables individuals to understand and predict the mental states of themselves and others~\citep{ToM, Premack1978DoesTC, Hurley2008TheSC, DBLP:journals/corr/abs-2404-13627}, has been used in games with incomplete information to anticipate opponents' actions and improve decision-making~\citep{DBLP:journals/ai/WeerdVV13}. Recent studies have investigated the ToM capabilities of large language models (LLMs) using text-based evaluations~\citep{DBLP:journals/corr/abs-2302-02083, DBLP:journals/corr/abs-2304-11490}. However, LLMs struggle with complex ToM inferences involving communication or second-order beliefs~\citep{ullman2023large}. The current study examines ToM within an interactive team task, where agents' mental states evolve dynamically, making the reasoning process more demanding. ToM has been applied to improve artificial agents' performance in various settings, such as integrating Bayesian Theory of Mind with optimal-planning agents~\citep{DBLP:conf/cogsci/LimTO20}, using SymbolicToM for reading comprehension tasks~\citep{DBLP:conf/acl/SclarKWS0T23}, and enhancing collaboration in multi-agent reinforcement learning~\citep{DBLP:journals/corr/abs-2307-01158, DBLP:journals/corr/abs-2110-00121}. Inspired by these studies, the present research aims to improve LLM-based agents' collaborative behaviors by incorporating explicit belief representations.

\begin{figure}[t]
  \vspace{-1.3cm}
  \centering
  \includegraphics[width=1\textwidth]{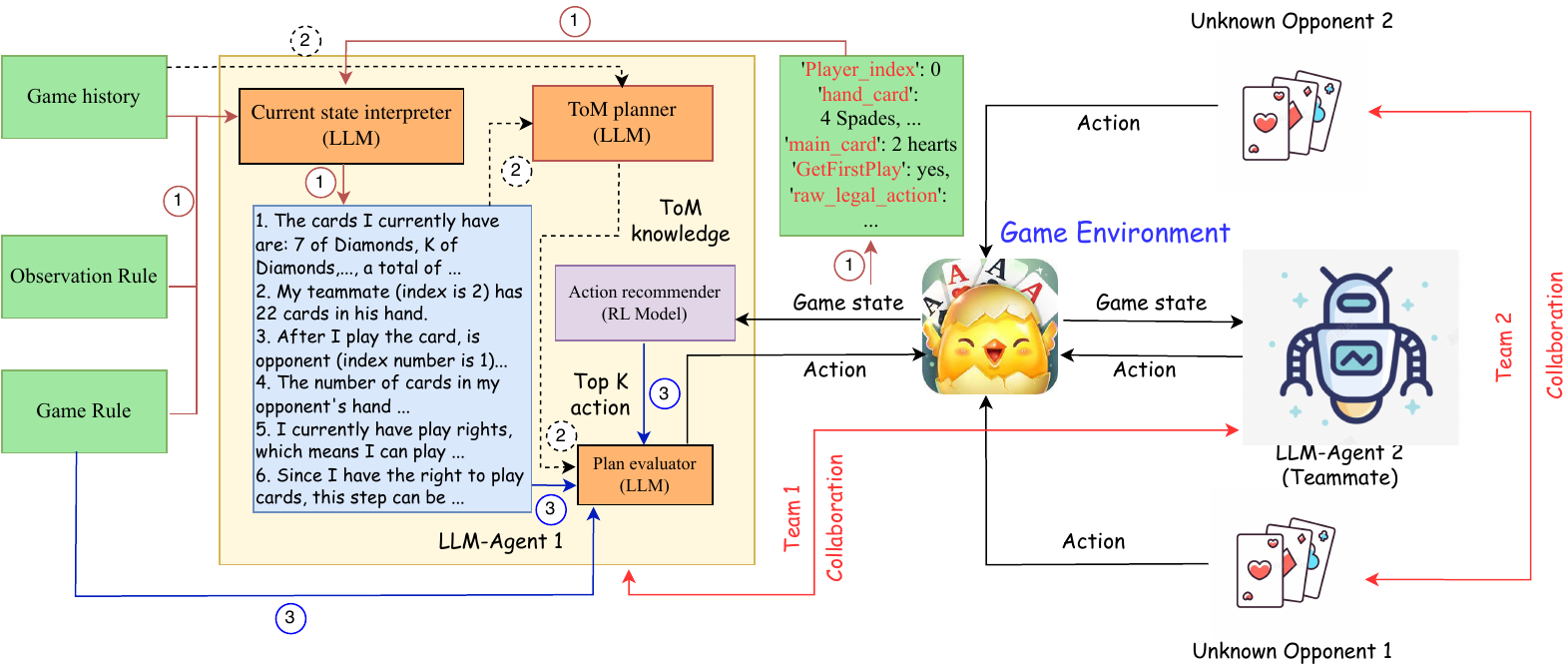}
  \vspace{-0.8cm}
  \caption{The figure illustrates the LLM-Agent's operational process, with the \textcolor{green!50!black}{green} square representing the model's input, the \textcolor{orange}{orange} square denoting the LLM model, the \textcolor{blue}{blue} square signifying the LLM model's output, and the \textcolor{purple}{purple} square corresponding to the RL-based model, which serves as an external tool to augment the agent's decision-making abilities.}
  \label{figure2}
  \vspace{-0.3cm}
\end{figure}

\section{LLM-based Emboddied agent for collaboration}

This study proposes a modular approach that enables LLM-based embodied agents to collaborate and strategically interact with adversarial agents in imperfect information games, Guandan, without requiring specialized training in the Chinese textual environment. The task is decomposed into essential components (Figure~\ref{figure2}), which include a State Interpreter, ToM Planner, and Plan Evaluator. The LLMs receive the game rules, observation rules (i.e., a prompt to guide LLM to convert low-level game state information to readable text), and game history as context. We provide a detailed description of the process of crafting prompts that guide LLMs to utilize their knowledge, reasoning abilities, and ToM to perform these modular functions and effectively navigate the complexities inherent in imperfect information games. 

\subsection{Game Rule, Observation Rule, and Game History}

This part refers to step 1 in Figure~\ref{figure2}. Inspired by~\cite{Suspicion-Agent}, we designed structured prompts to help LLMs understand the game rules and current state information, including card types, scoring rules, and win/loss conditions. The \textbf{game rule} template is shown as follows: 

\begin{itemize}
    \item\textbf{ General Rules:} A brief game introduction, team position rules, all single cards, and cards ranking.
    \item \textbf{Card Type that cannot beat other types:} \{Description of Card Type 1, Example of Card Type 1\}, \{Description of Card Type 2, Example of Card Type 2\}, ...;
    \item \textbf{Card Type that can beat other types:} \{Description of Card Type 1, Example of Card Type 1\}, \{Description of Card Type 2, Example of Card Type 2\}, ...;
    \item \textbf{Single Win/Loss Rule:} The scoring rules for four players in a single game, with different combinations of cards being played out in different order.
    \item \textbf{Whole Win/Loss Rule:} The overall win/loss conditions.
\end{itemize}

A sample template can be found in~\ref{app:a2}.

Several works~\citep{RLCard, Suspicion-Agent} note that game states are represented using low-level numerical values in most imperfect information game environments. LLMs can convert these atomic game states into human-readable text~\citep{DBLP:journals/corr/abs-2305-15486, DBLP:journals/corr/abs-2305-16291, DBLP:journals/corr/abs-2212-10846, Suspicion-Agent}, enhancing the model's understanding. Therefore, we structure the \textbf{observation rule} and \textbf{game history} conversion rules similarly to game rules to help LLM convert low-level current state and history information to readable text. The template includes input explanations, element descriptions, and conversion tips is shown as follows:

\begin{itemize}
    \item \textbf{Input Explanation:} The input types, like dictionaries and lists, are clearly specified. A description of every component in the input is also provided.
    \item \textbf{Conversion Tips:} Additional instructions guide converting the low-level game state representations into natural language descriptions.
\end{itemize}
An example of converted results can be found in~\ref{app:c1} and~\ref{app:c2}.

We can effectively convert low-level game states and history records into human-readable text, denoted as $\textit{Obs}_r$ and $\textit{His}_r$, respectively, by employing the game rule, observation conversion rule, and history conversion rule. The conditional distribution for each element $\textit{Obs}_r[j]$ within the generated text can be modeled using the prompts $\textit{Prompt}_{obs}$ as: $Obs_r \sim \prod_{j=1}^{N} LM_{\theta}(Obs_r[j]|Prompt_{obs}, Rule, Rule_{obs}, Obs_r[1,...,j-1])$, where $\textit{LM}_{\theta}$ represents the language model parameterized by $\theta$, and $\textit{N}$ is the length of the generated text $\textit{Obs}_r$. This \textbf{Current State Interpreter} module facilitates more intuitive interaction with the model in games with incomplete information.

\subsection{Tool use and Vanilla Planning Module}

In Guandan, the number of valid actions per turn can vary significantly, making it challenging for the model to accurately evaluate each action's merits. To address this, we incorporated an \textbf{action recommender}, a RL-based model. The model scores all valid actions available to the player in each round, allowing the LLM agent to focus on the top k highest score actions. We use the learned embedding from Danzero~\citep{lu2023danzero} to score every valid action.

This part refers to step 3 in Figure~\ref{figure2}. After transforming game states and history into a readable format and getting the top k actions selected by the action recommender, we design prompts to guide LLMs to find the best ways to play the game. We propose a vanilla planning approach that evaluates and predicts the consequences of executing each top k action and introduces a \textbf{plan evaluator} to make final decisions; a sample prompt template can be found in~\ref{app:c}.

\paragraph{Planning} Using the readable current observation, game history, game rules, and valid actions picked by the action recommender, we prompt LLMs to create text-based plans: 

\begin{equation}
O_{\textit{plan}} \sim \prod_{i=1}^{M} LM_{\theta}(O_{\textit{plan}}[i] | \textit{Prompt}_{\textit{plan}}, \textit{Rule}, Obs_{\textit{r}}, \textit{His}_r,  \textit{Top}_{k}action[i], O_{\textit{plan}}[1, ..., i-1])
\end{equation}

$\textit{i.e.}, O_{\textit{plan}} \sim F_{\theta}^{\textit{plans}}.$
The vanilla planning approach assumes a uniform marginal distribution for adversaries' and allies' actions, and it can be regarded as a zero-order ToM, denoted as $F_{\theta}^{\textit{zero-plan}}$.

\paragraph{Evaluator} To evaluate each plan's potential success, the \textbf{plan evaluator} will predict win rates for each plan with readable context $Obs_{\textit{r}}, \textit{His}_r$, game rule and then recommend the best action: 
\begin{equation}
\textit{a}_b = F^{zero-eval}(Obs_{r}, \textit{Rule}, O_{plan}, \textit{His}_r).
\end{equation}

\subsection{Planning with Theory of Mind (ToM) and Collaboration}

The part refers to the step 2 in Figure~\ref{figure2}. Vanilla planning techniques often struggle with ambiguities in games with imperfect information, especially when playing against opponents skilled at exploiting others' strategies. To address this, we developed a planning approach that employs the Theory of Mind (ToM) capabilities of LLMs \citep{ToM, DBLP:journals/corr/abs-2302-02083, Suspicion-Agent}. By understanding the actions of all opponents and teammates, this method enables the adaptation of strategies, enhancing success in complex, multi-agent game environments. We employ LLMs to analyze other agents' behavior and predict their actions using different ToM orders.

\textbf{Planning with First Order ToM: } In the first-order ToM modeling approach, LLM-Agent infers the card types that both opponents and teammates may possess based on their past actions and analyzes their strategies. This information is used to formulate confrontation or cooperative strategies. By inputting the game history $D_i = (h_1, h_2, \ldots, h_8)$, current observation $\textit{Obs}_r$, and game rules into LLMs, we prompt the model to examine the behavioral patterns exhibited by all agents: 

\begin{equation}
    \displaystyle{\textit{Belief} \sim \prod_{i=1}^{M} L_{\theta}(\textit{Belief}[i] | \textit{Prompt}_{\textit{pattern}}, \textit{Rule}, D^i,  \textit{Obs}_r, \textit{Belief}[1, \ldots, i-1])}.
\end{equation}
A sample prompt template for first-order behavior pattern analysis can be found in~\ref{app:b1}.

The predictions from the previous step can be used to enhance the \textbf{Planning} and \textbf{Evaluator} modules by incorporating $\textit{Belief}$ as an additional input. To optimize costs, we merge the \textbf{Planning} and \textbf{Evaluator} modules. A sample prompt template for the new Planning and
Evaluator modules can be found in~\ref{app:b2}, and the sample outputs of this part can be found in~\ref{app:c}.

\textbf{Planning with Second Order ToM: }
Skilled card game players excel at dynamically adapting their tactics, potentially tricking opponents into making suboptimal plays. Relying solely on first-order theoretical models can lead to misjudgments and gaps in the agent's cooperative approach. To address this, we introduce a planning method incorporating a second-order ToM. This enables the LLM agent to engage in more sophisticated reasoning by considering what the adversary believes about the LLM agents' intentions. This allows the agent to analyze the situation more comprehensively, mirroring the thought processes of high-level players. We include a simple prompt shown in~\ref{app:b3}. This method allows us to leverage Planning with a second-order Theory of Mind, enabling well-informed decisions and adaptive tactics based on the changing dynamics of the game. The sample outputs of this part we include in~\ref{app:c4}.

\begin{figure}[t]
    \vspace{-1.3cm}
    \centering
    \includegraphics[width=0.8\textwidth]{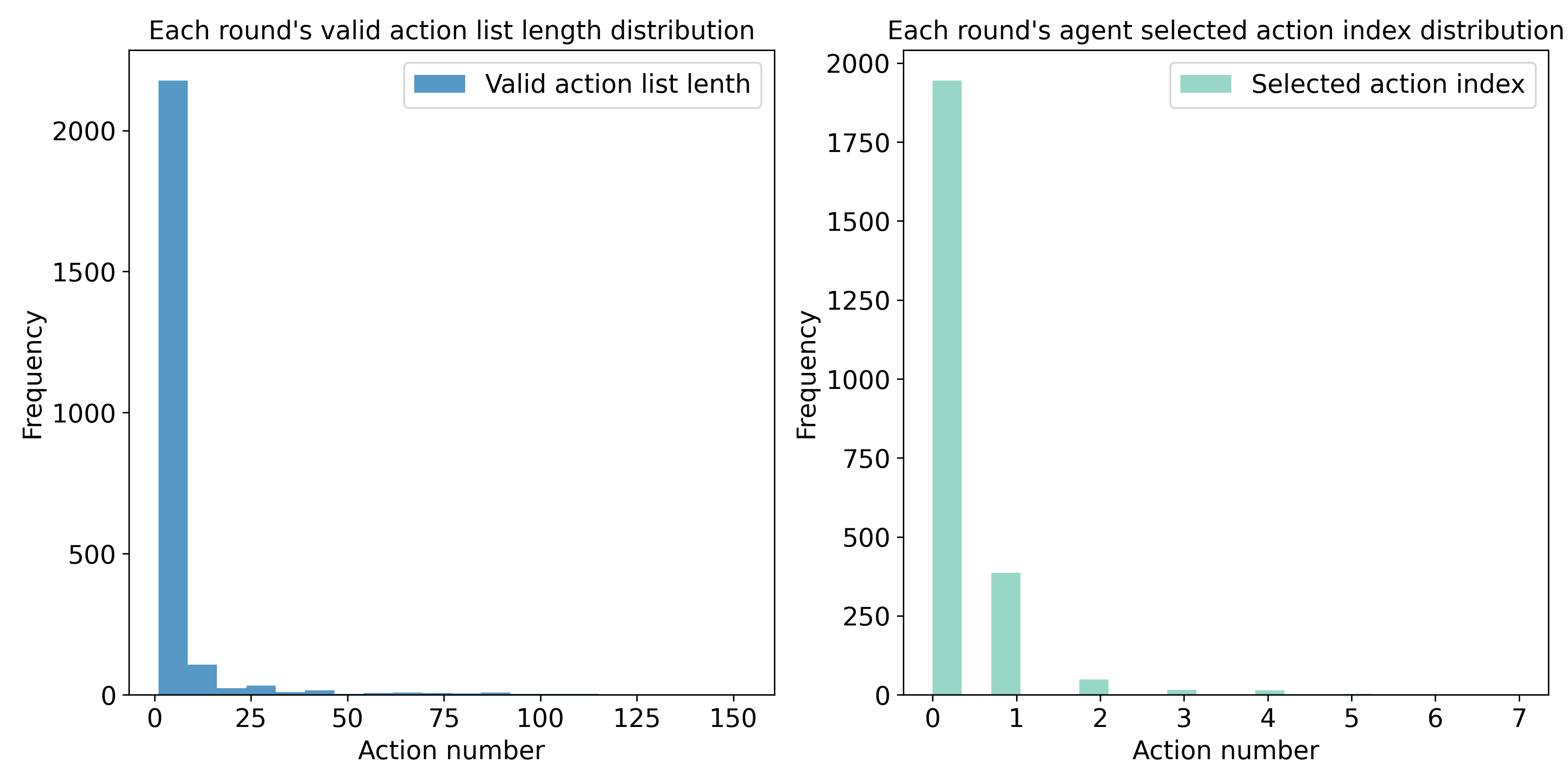}
    \vspace{-0.5cm}
    \caption{\textbf{Left Figure:} Distribution of agent's valid action list length per round. \textbf{Right Figure:} The distribution of agent's final selected action index. Results were obtained from about 2000 rounds.}
    \label{fig:histogram}
    \vspace{-0.3cm}
\end{figure}

\section{Experiment and Results}

\paragraph{Environments} To investigate the performance of LLM-based agents in a more realistic setting characterized by imperfect information, absence of communication, and collaborative dynamics, we selected Guandan as our experimental environment. We conducted tests on both open-source and close-source language models, including OpenAI's commercial model GPT-4-Turbo and GPT-3.5-Turbo~\citep{GPT4-report}, and Chinese-language open-source models like Baichuan2-7B-Chat, Baichuan2-13B-Chat, chatglm3-6b-32k, Qwen1.5-7B-Chat, and Qwen1.5-14B-Chat~\citep{baichuan2, chatglm, qwen}. In the test, each type of LLM agent will team up to compete against enemy teams consisting of other agent types. All experiments were conducted in a Chinese environment. The detailed rules and mechanics of Guandan are elaborated upon in Appendix \ref{app:a}.

\paragraph{Competing Methods}
\textbf{Danzero+}~\citep{lu2023danzero}: it is a state-of-the-art model and serves as the upper bound compared with other LLMs. Danzero+ uses the distributed framework to train the reinforcement learning model with the deep Monte Carlo method. They demonstrated their ability to perform at a human level. \textbf{Rule-based Agent:} selecting optimal card combinations under two distinct strategies: 'bestP\_lowV' and 'random'. The 'bestP\_lowV' strategy selects the best combination based on non-value criteria and calculated priority, maximizing strategic advantage while minimizing potential loss. It chooses the combination with the highest strategic priority and the lowest value among those with the optimal pattern. The 'random' strategy will execute a random action from the available options.

\begin{table}[t]
    \vspace{-1.3cm}
    \small
    \centering
    \scalebox{0.9}{\begin{tabular}{lcccc}
    \toprule
    & \multicolumn{3}{c}{Opponent Model} \\
        \cmidrule(lr){2-4}
        & \multirow{2}{*}{Random} & \multirow{2}{*}{Based-rule} & \multirow{2}{*}{DanZero+} & \multirow{2}{*}{Avg.} \\
        &  &  &  &  \\
    
        \midrule
        GPT-3.5-turbo (Vanilla) & +3.04 & +2.40 & -3.40 & +0.68 \\
        GPT-3.5-turbo (1st-ToM) & +3.20 & +3.00 & -3.27 & +0.98 \\
        GPT-3.5-turbo (2nd-ToM) & +3.00 & +2.90 & -3.00 & +0.97 \\
        
        GPT-4-turbo (Vanilla) & +3.78 & +2.98 & -2.97 & +1.26 \\
        GPT-4-turbo (1st-ToM) & +4.00 & +3.32 & -3.14 & +1.39 \\
        GPT-4-turbo (2nd-ToM) & +4.00 & +3.40 & -0.88 & +2.17 \\
    
        \midrule
        Baichuan2-Chat (7B) (Vanilla) & +3.00 & +1.90 & -4.00 & +0.30 \\
        Baichuan2-Chat (7B) (1st-ToM) & +3.30 & +3.00 & -4.00 & +0.76 \\
        Baichuan2-Chat (7B) (2nd-ToM) & +3.80 & +2.00 & -3.00 & +0.93 \\
        Baichuan2-Chat (13B) (Vanilla) & +3.67 & +3.50 & -3.80 & +1.12 \\
        Baichuan2-Chat (13B) (1st-ToM) & +3.60 & +2.25 & -3.30 & +0.85 \\
        Baichuan2-Chat (13B) (2nd-ToM) & +4.00 & +2.00 & -3.18 & +0.94 \\
        
        \midrule
        Chatglm3-32k (6B) (Vanilla) & +0.60 & -0.80 & -4.00 & -1.40 \\
        Chatglm3-32k (6B) (1st-ToM) & +1.80 & -0.40 & -3.40 & -0.67 \\
        Chatglm3-32k (6B) (2nd-ToM) & +1.40 & -0.20 & -3.00 & -0.60 \\
    
        \midrule
        Qwen1.5-Chat (7B) (Vanilla) & +3.10 & +2.40 & -4.00 & +0.5 \\
        Qwen1.5-Chat (7B) (1st-ToM) & +3.60 & +3.10 & -3.30 & +1.13 \\
        Qwen1.5-Chat (7B) (2nd-ToM) & +3.60 & +3.00 & -2.80 & +1.27 \\
        Qwen1.5-Chat (14B) (Vanilla) & +3.15 & +3.20 & -4.00 & +0.78 \\
        Qwen1.5-Chat (14B) (1st-ToM) & +3.20 & +3.40 & -3.62 & +0.99 \\
        Qwen1.5-Chat (14B) (2nd-ToM) & +3.67 & +3.40 & -2.98 & +1.36 \\
        
        \bottomrule
    \end{tabular}}
    \vspace{-0.3cm}
    \caption{The average performance of various LLMs against random, rule-based, and Danzero+ opponents in all game rounds.     
    For the game scoring rule, if a team's agents are the first and second to play out their cards, they score 4 points, while the other team loses 4 points, more details are in Appendix~\ref{app:a1}. The first three columns show the LLM agent's average score in 40 games against other agents. The last column is the LLM agent's average performance against three agents.
    }
    \label{tab:table1}
    \vspace{-0.2cm}
\end{table}

\begin{table}[t]
    \small
    \centering
    \resizebox{\textwidth}{!}{%
    \scalebox{0.9}{\begin{tabular}{lccccccc}
    \toprule
        & \multicolumn{6}{c}{Opponent Model} \\
        \cmidrule(lr){2-7}
        & Random & Random & DanZero+ & DanZero+ & Ours & Ours & \multirow{2}{*}{Avg.} \\
        & (Pos 0 \& 2) & (Pos 1 \& 3) & (Pos 0 \& 2) & (Pos 1 \& 3) & (Pos 0 \& 2) & (Pos 1 \& 3) & \\
        
        \midrule
        Random & - & - & -4.00 & -4.00 & -3.36 & -2.78 & -3.53 \\
        DanZero+ & +4.00 & +4.00 & - & - & +2.50 & +4.00 & 3.63\\
        Ours (GPT-3.5) & +2.78 & +3.36 & -4.00 & -2.50 & - & - & -0.12\\
        \bottomrule
    \end{tabular}}
    }
    \vspace{-0.3cm}
    \caption{The table compares GPT-3.5's performance when playing with Random and Danzero+ in different positions in Guandan environments over 30 games, with 15 rounds played in each of the two positions.}
    \label{tab:table2}
\end{table}

\paragraph{Evaluation Methods} We follow~\cite{Suspicion-Agent} to design a dual-method evaluation framework to ensure a more robust assessment of the LLM-agent's performance, addressing potential factors that may provide inherent advantages in this type of game. \textbf{Variable random seeding: } The method involves the investigated agents playing 40 games against different baselines, each utilizing a unique random seed to mitigate variability arising from random seed setting; results are shown in Table~\ref{tab:table1}. \textbf{Position swapping: } This method focuses on the potential influence of a player's position on the game's outcome, as it often determines the situation in the Guandan game. The investigated agent was initially assigned positions 0 and 2 for the first 20 games. The positions of the investigated agent and the baseline model were then exchanged for a repeat of the 20 games. This approach allows for assessing the impact of position on the game's winning rate, with the results shown in Table~\ref{tab:table2}.

\paragraph{Action distribution} Since Guandan offers a dynamic length of valid actions list for each agent to choose from in every round. We analyzed 10 games, recording the number of valid actions available to LLM-Agent whenever it was their turn to play a card, along with the action index they ultimately selected (totaling approximately 1900 actions). Notably, although LLM-Agent often had more than 20 possible actions to select from during their turn, their chosen actions were consistently restricted to a maximum of 7. Moreover, they strongly preferred the first five actions, as illustrated in Figure~\ref{fig:histogram}. We believe that when an LLM faces too many legal actions, it often has trouble effectively processing the extensive text input. This leads to a reduced understanding of each action, especially those action elements at the end of the valid action list. Moreover, when there are many valid actions, it becomes impractical for the model to generate and analyze a complete set of solutions for each action. To overcome these limitations and improve the model's performance, we introduced the action recommender tool as a crucial part of our approach.

\paragraph{Results Analysis} (1) \textbf{ToM planning method outperforms Vanilla planning method: } Table~\ref{tab:table1} shows that injecting belief knowledge obtained through ToM planning consistently improves the performance of almost all models. This highlights the importance of ToM knowledge in playing card games. Moreover, it is clear that LLM-agent has a natural ability to analyze other players' beliefs, and developing this skill dramatically contributes to its overall performance.
(2) \textbf{The gap between the GPT-4 and other models: } Although ToM capability improves nearly all models, GPT-4 consistently outperforms other LLMs. This superiority is particularly evident when the agents are pitted against the three baselines. In these matchups, the agent based on GPT-4 achieves an impressive average score of +2.17, demonstrating its exceptional skills in the Guandan card game. In contrast, the best performance among the other LLMs reaches only +1.36, a notable difference of 0.81 points compared to GPT-4. This disparity in performance can be attributed to GPT-4's advanced ability to leverage ToM knowledge better and make more strategic decisions in the complex environment of Guandan. (3) \textbf{Danzero+ outperforms all LLM-based agents: } The findings indicate that none of the LLM-based embodied agents, with the exception of the GPT-4 model, can outperform the current state-of-the-art model, danzero+, in this game. Nonetheless, the GPT-4 model exhibits an impressive capability to come close to danzero+'s performance, attaining a score of -0.88 with the second ToM planning method, which is remarkably near. This observation highlights the significant limitations of LLMs without fine-tuning when faced with cooperative tasks involving imperfect information, as most of these models struggle to address such challenges effectively. (4) \textbf{Position impact: } In this experimental segment, we employ GPT-3.5 as the evaluation model. The results shown in Table~\ref{tab:table2} indicate that when the GPT-3.5 model competes against the random and Danzero+ models, it achieves a significantly higher win rate when positioned at 0 and 2. Notably, when facing danzero+, the model secures -2.5 points in positions 0 and 2, whereas changing positions yields only -4 points. This observation aligns with our perspective, as being in the 0 \& 2 positions grants the LLM-agent team the advantage of playing cards first. In card games, the right to initiate play often influences the game's outcome, and GPT-3.5 demonstrates a superior ability to capitalize on this advantage.

\subsection{Ablation Study and Component Analysis}

\paragraph{Ablation Study on Different Languages: }Guandan is widely popular in China. Due to its cultural specificity, many game-related terms take time to translate accurately into other languages. To investigate the impact of different language environments on the performance of LLM agents, we conducted GPT-3.5 as a testbed model. Table~\ref{tab:table3} presents the performance of GPT-3.5 with various ToM planning methods in both Chinese and English game environments. The results demonstrate that the LLM-Agent's performance in the Chinese language environment significantly surpasses its performance in the English language environment. This phenomenon can be attributed to the fact that GPT-3.5 is pre-trained on an enormous amount of data, and most Guandan-related data is primarily available in Chinese. Consequently, we opted to use Chinese as the language environment for our experiments to leverage the language model's inherent knowledge and maximize its performance in Guandan. 

\paragraph{Ablation Study on the Action Recommender: } As illustrated in Figure~\ref{fig:histogram}, we observed that the actions selected by the LLM-Agent are primarily concentrated within the first five actions, even when presented with more than 50 options. To address this limitation, we introduced an action recommender component, an external tool, specifically an RL-based model, to assign scores to each available action and enable the LLM-Agent to invoke it. This tool returns the top 5 actions with the highest scores by default. In this section, we primarily investigate the impact of the action recommender component on the LLM-Agent's performance. We compare three scenarios: (1) the absence of an action recommender, (2) the inclusion of an action recommender returning the top 5 actions, and (3) the inclusion of an action recommender returning the top 3 actions. Furthermore, we conduct these comparisons using GPT-3.5 with various planning methods to assess the effectiveness of the action recommender across different configurations. 

\begin{table}[t]
    \vspace{-1cm}
    \centering
    \small
    \begin{tabular}{lcccc}
    \toprule
    & \multicolumn{3}{c}{Opponent Model} \\
    \cmidrule(lr){2-4}
    & \multirow{2}{*}{Random} & \multirow{2}{*}{Based-rule} & \multirow{2}{*}{DanZero+} & \multirow{2}{*}{Avg.} \\
    &  &  &  &  \\

    Ours (Vanilla) (English) & +1.00 & -3.00 & -4.00 & -2.00 \\
    Ours (1st-ToM) (English) & +2.60 & -0.60 & -4.00 & -0.67 \\
    Ours (2nd-ToM) (English) & +2.40 & -0.8 & -4.00 & -0.80 \\
    
    \midrule
    Ours (Vanilla) (Chinese)& +3.04 & +2.40 & -3.40 & +0.68 \\
    Ours (1st-ToM) (Chinese)& +3.20 & +3.00 & -3.27 & +0.98 \\
    Ours (2nd-ToM) (Chinese)& +3.00 & +2.90 & -3.00 & +0.97 \\
    
    \bottomrule
    \end{tabular}
    \caption{This table compares the LLM-Agent's performance across variations of language environments in 30 games using GPT-3.5 with TopK=5.}
    \label{tab:table3}
\end{table}

\begin{table}[t]
    \small
    \centering
    \begin{tabular}{lcccc}
    \toprule
    & \multicolumn{3}{c}{Opponent Model} \\
    \cmidrule(lr){2-4}
    & \multirow{2}{*}{Random} & \multirow{2}{*}{Based-rule} & \multirow{2}{*}{DanZero+} & \multirow{2}{*}{Avg.} \\
    &  &  &  &  \\

    Ours (w/o Tool) (Vanilla) & -0.60 & -1.80 & -4.00 & -2.13 \\
    Ours (w/o Tool) (1st-ToM) & +2.24 & -3.00 & -4.00 & -1.59 \\
    Ours (w/o Tool) (2nd-ToM) & +1.80 & -2.57 & -4.00 & -1.59 \\
    
    \midrule
    Ours (Top k = 3) (Vanilla) & +2.88 & +1.62 & -3.21 & +0.43 \\
    Ours (Top k = 3) (1st-ToM) & +2.96 & +1.50 & -3.42 & +0.35 \\
    Ours (Top k = 3) (2nd-ToM) & +2.82 & +2.30 & -3.02 & +0.70 \\

    \midrule
    Ours (Top k = 5) (Vanilla) & +3.04 & +2.40 & -3.40 & +0.68 \\
    Ours (Top k = 5) (1st-ToM) & +3.20 & +3.00 & -3.27 & +0.98 \\
    Ours (Top k = 5) (2nd-ToM) & +3.00 & +2.90 & -3.00 & +0.97 \\
    
    \bottomrule
\end{tabular}
    \caption{The comparison results of including an action recommender or not, and the effect of different top-k actions on results, using GPT-3.5 tested on 30 games.}
    \label{tab:table4}
\end{table}

The results shown in Table~\ref{tab:table4} allow for the following observation: (1) \textbf{LLM capabilities can be improved with the help of external tools: }The utilization of this external tool significantly improves the performance of the LLM-Agent in this game, which demonstrates that LLM-Agents struggle with long reasoning problems. Even when the game environment provides the agent with a relatively extensive list of valid actions, the agent cannot effectively process such lengthy text and consequently focuses only on the first few actions. The introduction of the action recommender, which only returns the top k operations, alleviates this problem and enhances the agent's capabilities. (2) \textbf{The gap between different numbers of top k action: } The results demonstrate that setting K to 5 leads to superior model performance compared to K=3. This finding indicates that the RL-based tool may not always produce optimal results. When the LLM-Agent is restricted to choosing from only three actions, it may not be able to fully leverage its capabilities, especially when significantly more than three valid actions are available in a given scenario. In such cases, the model is constrained to analyzing and selecting from a limited subset of possibilities. Additionally, the improvement in the ToM planning method is less significant when K=3 compared to K =5. This observation further supports the idea that the model's ToM capabilities cannot be fully utilized when K is too small.

\section{Conclusion}

This study compares open-source and closed-source LLMs with SOTA-RL-based models in emulating opponent and teammate behavior in the complex card game Guandan. LLMs exhibit subpar performance compared to RL-based approaches in intricate and realistic scenarios. We propose a ToM planning method to optimize LLM-agent's decision-making and collaboration in multi-agent gaming environments. We also develop an RL-based model to address LLMs' challenges in adapting to dynamic changes and extensive legal action lists. This research deepens our understanding of LLMs' potential in navigating complex social interactions and advances cognitive modeling in AI.

\section*{Ethics Statement}
In this study, we conducted experiments and evaluations on large language models (LLMs) for their applicability and performance in the Guandan card game, imperfect information, and multi-agent collaboration tasks. We compared open-source and closed-source LLMs to ensure a fair and transparent evaluation, aiming to contribute to the broader understanding of LLMs and their potential to address complex tasks without bias. Our research adheres to the highest ethical standards, ensuring that the data used does not contain personally identifiable information and focuses solely on the performance of LLMs in the context of the Guandan card game. While employing the Theory of Mind (ToM) concept, we do not attribute human-like consciousness to these AI models. Our objective is to investigate the potential of LLMs in complex social interactions and strategic thinking, acknowledging the potential risks and emphasizing the importance of responsible AI development, deployment, and use.

\section*{Reproducibility Statement}
To ensure the reproducibility of our research, we have taken several steps. We have comprehensively described our methods, including the planning approaches, the action recommender, and the ToM-aware LLMs. This detailed account of our methodology enables other researchers to replicate our experiments and build upon our findings. We have evaluated open-source and closed-source LLMs, ensuring that researchers with access to either model type can reproduce our results. Furthermore, to facilitate further research and promote transparency, we have committed to making all prompts and codes publicly available in our codebase. This will enable other researchers to reproduce our experiments and verify our results. By taking these steps, we aim to ensure that our research is transparent and reproducible and contributes to the broader understanding of LLMs and their potential in addressing complex, real-world tasks.





\bibliography{custom}

\begin{thebibliography}{59}
\providecommand{\natexlab}[1]{#1}
\providecommand{\url}[1]{\texttt{#1}}
\expandafter\ifx\csname urlstyle\endcsname\relax
  \providecommand{\doi}[1]{doi: #1}\else
  \providecommand{\doi}{doi: \begingroup \urlstyle{rm}\Url}\fi

\bibitem[Agashe et~al.(2023)Agashe, Fan, and Wang]{DBLP:journals/corr/abs-2310-03903}
Saaket Agashe, Yue Fan, and Xin~Eric Wang.
\newblock Evaluating multi-agent coordination abilities in large language models.
\newblock \emph{CoRR}, abs/2310.03903, 2023.
\newblock \doi{10.48550/ARXIV.2310.03903}.
\newblock URL \url{https://doi.org/10.48550/arXiv.2310.03903}.

\bibitem[Bai et~al.(2023)Bai, Bai, Chu, Cui, Dang, Deng, Fan, Ge, Han, Huang, Hui, Ji, Li, Lin, Lin, Liu, Liu, Lu, Lu, Ma, Men, Ren, Ren, Tan, Tan, Tu, Wang, Wang, Wang, Wu, Xu, Xu, Yang, Yang, Yang, Yang, Yao, Yu, Yuan, Yuan, Zhang, Zhang, Zhang, Zhang, Zhou, Zhou, Zhou, and Zhu]{qwen}
Jinze Bai, Shuai Bai, Yunfei Chu, Zeyu Cui, Kai Dang, Xiaodong Deng, Yang Fan, Wenbin Ge, Yu~Han, Fei Huang, Binyuan Hui, Luo Ji, Mei Li, Junyang Lin, Runji Lin, Dayiheng Liu, Gao Liu, Chengqiang Lu, Keming Lu, Jianxin Ma, Rui Men, Xingzhang Ren, Xuancheng Ren, Chuanqi Tan, Sinan Tan, Jianhong Tu, Peng Wang, Shijie Wang, Wei Wang, Shengguang Wu, Benfeng Xu, Jin Xu, An~Yang, Hao Yang, Jian Yang, Shusheng Yang, Yang Yao, Bowen Yu, Hongyi Yuan, Zheng Yuan, Jianwei Zhang, Xingxuan Zhang, Yichang Zhang, Zhenru Zhang, Chang Zhou, Jingren Zhou, Xiaohuan Zhou, and Tianhang Zhu.
\newblock Qwen technical report.
\newblock \emph{CoRR}, abs/2309.16609, 2023.
\newblock \doi{10.48550/ARXIV.2309.16609}.
\newblock URL \url{https://doi.org/10.48550/arXiv.2309.16609}.

\bibitem[Bakhtin et~al.(2022)Bakhtin, Brown, Dinan, Farina, Flaherty, Fried, Goff, Gray, Hu, Jacob, Komeili, Konath, Kwon, Lerer, Lewis, Miller, Mitts, Renduchintala, Roller, Rowe, Shi, Spisak, Wei, Wu, Zhang, and Zijlstra]{Bakhtin2022HumanlevelPI}
Anton Bakhtin, Noam Brown, Emily Dinan, Gabriele Farina, Colin Flaherty, Daniel Fried, Andrew Goff, Jonathan Gray, Hengyuan Hu, Athul~Paul Jacob, Mojtaba Komeili, Karthik Konath, Minae Kwon, Adam Lerer, Mike Lewis, Alexander~H. Miller, Sandra Mitts, Adithya Renduchintala, Stephen Roller, Dirk Rowe, Weiyan Shi, Joe Spisak, Alexander Wei, David~J. Wu, Hugh Zhang, and Markus Zijlstra.
\newblock Human-level play in the game of diplomacy by combining language models with strategic reasoning.
\newblock \emph{Science}, 378:\penalty0 1067 -- 1074, 2022.
\newblock URL \url{https://api.semanticscholar.org/CorpusID:253759631}.

\bibitem[Brown \& Sandholm(2018)Brown and Sandholm]{Brown2018SuperhumanAF}
Noam Brown and Tuomas Sandholm.
\newblock Superhuman ai for heads-up no-limit poker: Libratus beats top professionals.
\newblock \emph{Science}, 359:\penalty0 418 -- 424, 2018.
\newblock URL \url{https://api.semanticscholar.org/CorpusID:5003977}.

\bibitem[Bubeck et~al.(2023)Bubeck, Chandrasekaran, Eldan, Gehrke, Horvitz, Kamar, Lee, Lee, Li, Lundberg, Nori, Palangi, Ribeiro, and Zhang]{DBLP:journals/corr/abs-2303-12712}
S{\'{e}}bastien Bubeck, Varun Chandrasekaran, Ronen Eldan, Johannes Gehrke, Eric Horvitz, Ece Kamar, Peter Lee, Yin~Tat Lee, Yuanzhi Li, Scott~M. Lundberg, Harsha Nori, Hamid Palangi, Marco~T{\'{u}}lio Ribeiro, and Yi~Zhang.
\newblock Sparks of artificial general intelligence: Early experiments with {GPT-4}.
\newblock \emph{CoRR}, abs/2303.12712, 2023.

\bibitem[Carroll et~al.(2019)Carroll, Shah, Ho, Griffiths, Seshia, Abbeel, and Dragan]{DBLP:conf/nips/CarrollSHGSAD19}
Micah Carroll, Rohin Shah, Mark~K. Ho, Tom Griffiths, Sanjit~A. Seshia, Pieter Abbeel, and Anca~D. Dragan.
\newblock On the utility of learning about humans for human-ai coordination.
\newblock In Hanna~M. Wallach, Hugo Larochelle, Alina Beygelzimer, Florence d'Alch{\'{e}}{-}Buc, Emily~B. Fox, and Roman Garnett (eds.), \emph{Advances in Neural Information Processing Systems 32: Annual Conference on Neural Information Processing Systems 2019, NeurIPS 2019, December 8-14, 2019, Vancouver, BC, Canada}, pp.\  5175--5186, 2019.
\newblock URL \url{https://proceedings.neurips.cc/paper/2019/hash/f5b1b89d98b7286673128a5fb112cb9a-Abstract.html}.

\bibitem[Chan et~al.(2023{\natexlab{a}})Chan, Liu, Chan, Cheng, Song, Wong, and See]{DBLP:journals/corr/abs-2309-08303}
Chunkit Chan, Xin Liu, Tsz~Ho Chan, Jiayang Cheng, Yangqiu Song, Ginny~Y. Wong, and Simon See.
\newblock Self-consistent narrative prompts on abductive natural language inference.
\newblock \emph{CoRR}, abs/2309.08303, 2023{\natexlab{a}}.
\newblock \doi{10.48550/arXiv.2309.08303}.
\newblock URL \url{https://doi.org/10.48550/arXiv.2309.08303}.

\bibitem[Chan et~al.(2023{\natexlab{b}})Chan, Liu, Cheng, Li, Song, Wong, and See]{DBLP:conf/acl/ChanLCLSWS23}
Chunkit Chan, Xin Liu, Jiayang Cheng, Zihan Li, Yangqiu Song, Ginny~Y. Wong, and Simon See.
\newblock Discoprompt: Path prediction prompt tuning for implicit discourse relation recognition.
\newblock In Anna Rogers, Jordan~L. Boyd{-}Graber, and Naoaki Okazaki (eds.), \emph{Findings of the Association for Computational Linguistics: {ACL} 2023, Toronto, Canada, July 9-14, 2023}, pp.\  35--57. Association for Computational Linguistics, 2023{\natexlab{b}}.
\newblock \doi{10.18653/v1/2023.findings-acl.4}.
\newblock URL \url{https://doi.org/10.18653/v1/2023.findings-acl.4}.

\bibitem[Chan et~al.(2024{\natexlab{a}})Chan, Cheng, Wang, Jiang, Fang, Liu, and Song]{DBLP:conf/eacl/ChanCWJFLS24}
Chunkit Chan, Jiayang Cheng, Weiqi Wang, Yuxin Jiang, Tianqing Fang, Xin Liu, and Yangqiu Song.
\newblock Exploring the potential of chatgpt on sentence level relations: {A} focus on temporal, causal, and discourse relations.
\newblock In Yvette Graham and Matthew Purver (eds.), \emph{Findings of the Association for Computational Linguistics: {EACL} 2024, St. Julian's, Malta, March 17-22, 2024}, pp.\  684--721. Association for Computational Linguistics, 2024{\natexlab{a}}.
\newblock URL \url{https://aclanthology.org/2024.findings-eacl.47}.

\bibitem[Chan et~al.(2024{\natexlab{b}})Chan, Jiayang, Yim, Deng, Fan, Li, Liu, Zhang, Wang, and Song]{DBLP:journals/corr/abs-2404-13627}
Chunkit Chan, Cheng Jiayang, Yauwai Yim, Zheye Deng, Wei Fan, Haoran Li, Xin Liu, Hongming Zhang, Weiqi Wang, and Yangqiu Song.
\newblock Negotiationtom: {A} benchmark for stress-testing machine theory of mind on negotiation surrounding.
\newblock \emph{CoRR}, abs/2404.13627, 2024{\natexlab{b}}.
\newblock \doi{10.48550/ARXIV.2404.13627}.
\newblock URL \url{https://doi.org/10.48550/arXiv.2404.13627}.

\bibitem[Cheng et~al.(2023)Cheng, Qiu, Chan, Fang, Wang, Chan, Ru, Guo, Zhang, Song, Zhang, and Zhang]{DBLP:conf/emnlp/ChengQCFWCRGZSZ23}
Jiayang Cheng, Lin Qiu, Tsz~Ho Chan, Tianqing Fang, Weiqi Wang, Chunkit Chan, Dongyu Ru, Qipeng Guo, Hongming Zhang, Yangqiu Song, Yue Zhang, and Zheng Zhang.
\newblock Storyanalogy: Deriving story-level analogies from large language models to unlock analogical understanding.
\newblock In Houda Bouamor, Juan Pino, and Kalika Bali (eds.), \emph{Proceedings of the 2023 Conference on Empirical Methods in Natural Language Processing, {EMNLP} 2023, Singapore, December 6-10, 2023}, pp.\  11518--11537. Association for Computational Linguistics, 2023.
\newblock \doi{10.18653/V1/2023.EMNLP-MAIN.706}.
\newblock URL \url{https://doi.org/10.18653/v1/2023.emnlp-main.706}.

\bibitem[Chiang et~al.(2023)Chiang, Li, Lin, Sheng, Wu, Zhang, Zheng, Zhuang, Zhuang, Gonzalez, Stoica, and Xing]{vicuna2023}
Wei-Lin Chiang, Zhuohan Li, Zi~Lin, Ying Sheng, Zhanghao Wu, Hao Zhang, Lianmin Zheng, Siyuan Zhuang, Yonghao Zhuang, Joseph~E. Gonzalez, Ion Stoica, and Eric~P. Xing.
\newblock Vicuna: An open-source chatbot impressing gpt-4 with 90\%* chatgpt quality, March 2023.
\newblock URL \url{https://lmsys.org/blog/2023-03-30-vicuna/}.

\bibitem[Chowdhery et~al.(2023)Chowdhery, Narang, Devlin, Bosma, Mishra, Roberts, Barham, Chung, Sutton, Gehrmann, Schuh, Shi, Tsvyashchenko, Maynez, Rao, Barnes, Tay, Shazeer, Prabhakaran, Reif, Du, Hutchinson, Pope, Bradbury, Austin, Isard, Gur{-}Ari, Yin, Duke, Levskaya, Ghemawat, Dev, Michalewski, Garcia, Misra, Robinson, Fedus, Zhou, Ippolito, Luan, Lim, Zoph, Spiridonov, Sepassi, Dohan, Agrawal, Omernick, Dai, Pillai, Pellat, Lewkowycz, Moreira, Child, Polozov, Lee, Zhou, Wang, Saeta, Diaz, Firat, Catasta, Wei, Meier{-}Hellstern, Eck, Dean, Petrov, and Fiedel]{DBLP:journals/jmlr/ChowdheryNDBMRBCSGSSTMRBTSPRDHPBAI23}
Aakanksha Chowdhery, Sharan Narang, Jacob Devlin, Maarten Bosma, Gaurav Mishra, Adam Roberts, Paul Barham, Hyung~Won Chung, Charles Sutton, Sebastian Gehrmann, Parker Schuh, Kensen Shi, Sasha Tsvyashchenko, Joshua Maynez, Abhishek Rao, Parker Barnes, Yi~Tay, Noam Shazeer, Vinodkumar Prabhakaran, Emily Reif, Nan Du, Ben Hutchinson, Reiner Pope, James Bradbury, Jacob Austin, Michael Isard, Guy Gur{-}Ari, Pengcheng Yin, Toju Duke, Anselm Levskaya, Sanjay Ghemawat, Sunipa Dev, Henryk Michalewski, Xavier Garcia, Vedant Misra, Kevin Robinson, Liam Fedus, Denny Zhou, Daphne Ippolito, David Luan, Hyeontaek Lim, Barret Zoph, Alexander Spiridonov, Ryan Sepassi, David Dohan, Shivani Agrawal, Mark Omernick, Andrew~M. Dai, Thanumalayan~Sankaranarayana Pillai, Marie Pellat, Aitor Lewkowycz, Erica Moreira, Rewon Child, Oleksandr Polozov, Katherine Lee, Zongwei Zhou, Xuezhi Wang, Brennan Saeta, Mark Diaz, Orhan Firat, Michele Catasta, Jason Wei, Kathy Meier{-}Hellstern, Douglas Eck, Jeff Dean, Slav Petrov, and Noah Fiedel.
\newblock Palm: Scaling language modeling with pathways.
\newblock \emph{J. Mach. Learn. Res.}, 24:\penalty0 240:1--240:113, 2023.
\newblock URL \url{http://jmlr.org/papers/v24/22-1144.html}.

\bibitem[de~Weerd et~al.(2013)de~Weerd, Verbrugge, and Verheij]{DBLP:journals/ai/WeerdVV13}
Harmen de~Weerd, Rineke Verbrugge, and Bart Verheij.
\newblock How much does it help to know what she knows you know? an agent-based simulation study.
\newblock \emph{Artif. Intell.}, 199-200:\penalty0 67--92, 2013.
\newblock \doi{10.1016/J.ARTINT.2013.05.004}.
\newblock URL \url{https://doi.org/10.1016/j.artint.2013.05.004}.

\bibitem[Deng et~al.(2024)Deng, Chan, Wang, Sun, Fan, Zheng, Yim, and Song]{DBLP:journals/corr/abs-2404-14215}
Zheye Deng, Chunkit Chan, Weiqi Wang, Yuxi Sun, Wei Fan, Tianshi Zheng, Yauwai Yim, and Yangqiu Song.
\newblock Text-tuple-table: Towards information integration in text-to-table generation via global tuple extraction.
\newblock \emph{CoRR}, abs/2404.14215, 2024.
\newblock \doi{10.48550/ARXIV.2404.14215}.
\newblock URL \url{https://doi.org/10.48550/arXiv.2404.14215}.

\bibitem[Frieder et~al.(2023)Frieder, Pinchetti, Griffiths, Salvatori, Lukasiewicz, Petersen, Chevalier, and Berner]{DBLP:journals/corr/abs-2301-13867}
Simon Frieder, Luca Pinchetti, Ryan{-}Rhys Griffiths, Tommaso Salvatori, Thomas Lukasiewicz, Philipp~Christian Petersen, Alexis Chevalier, and Julius Berner.
\newblock Mathematical capabilities of chatgpt.
\newblock \emph{CoRR}, abs/2301.13867, 2023.

\bibitem[Frith \& Frith(2005)Frith and Frith]{ToM}
Chris Frith and Uta Frith.
\newblock Theory of mind.
\newblock \emph{Curr Biol}, 15:\penalty0 R644--6, 01 2005.

\bibitem[Gray et~al.(2021)Gray, Lerer, Bakhtin, and Brown]{DBLP:conf/iclr/GrayLBB21}
Jonathan Gray, Adam Lerer, Anton Bakhtin, and Noam Brown.
\newblock Human-level performance in no-press diplomacy via equilibrium search.
\newblock In \emph{9th International Conference on Learning Representations, {ICLR} 2021, Virtual Event, Austria, May 3-7, 2021}. OpenReview.net, 2021.
\newblock URL \url{https://openreview.net/forum?id=0-uUGPbIjD}.

\bibitem[Guo et~al.(2022)Guo, Li, Li, Tiong, Li, Tao, and Hoi]{DBLP:journals/corr/abs-2212-10846}
Jiaxian Guo, Junnan Li, Dongxu Li, Anthony Meng~Huat Tiong, Boyang Li, Dacheng Tao, and Steven C.~H. Hoi.
\newblock From images to textual prompts: Zero-shot {VQA} with frozen large language models.
\newblock \emph{CoRR}, abs/2212.10846, 2022.
\newblock \doi{10.48550/ARXIV.2212.10846}.
\newblock URL \url{https://doi.org/10.48550/arXiv.2212.10846}.

\bibitem[Guo et~al.(2023)Guo, Yang, Yoo, Lin, Iwasawa, and Matsuo]{Suspicion-Agent}
Jiaxian Guo, Bo~Yang, Paul Yoo, Bill~Yuchen Lin, Yusuke Iwasawa, and Yutaka Matsuo.
\newblock Suspicion-agent: Playing imperfect information games with theory of mind aware {GPT-4}.
\newblock \emph{CoRR}, abs/2309.17277, 2023.
\newblock \doi{10.48550/ARXIV.2309.17277}.
\newblock URL \url{https://doi.org/10.48550/arXiv.2309.17277}.

\bibitem[Huang et~al.(2022)Huang, Abbeel, Pathak, and Mordatch]{DBLP:conf/icml/HuangAPM22}
Wenlong Huang, Pieter Abbeel, Deepak Pathak, and Igor Mordatch.
\newblock Language models as zero-shot planners: Extracting actionable knowledge for embodied agents.
\newblock In Kamalika Chaudhuri, Stefanie Jegelka, Le~Song, Csaba Szepesv{\'{a}}ri, Gang Niu, and Sivan Sabato (eds.), \emph{International Conference on Machine Learning, {ICML} 2022, 17-23 July 2022, Baltimore, Maryland, {USA}}, volume 162 of \emph{Proceedings of Machine Learning Research}, pp.\  9118--9147. {PMLR}, 2022.
\newblock URL \url{https://proceedings.mlr.press/v162/huang22a.html}.

\bibitem[Hurley(2008)]{Hurley2008TheSC}
Susan Hurley.
\newblock The shared circuits model (scm): how control, mirroring, and simulation can enable imitation, deliberation, and mindreading.
\newblock \emph{The Behavioral and brain sciences}, 31 1:\penalty0 1--22; discussion 22--58, 2008.
\newblock URL \url{https://api.semanticscholar.org/CorpusID:16427883}.

\bibitem[Jiang et~al.(2023)Jiang, Chan, Chen, and Wang]{DBLP:journals/corr/abs-2305-12870}
Yuxin Jiang, Chunkit Chan, Mingyang Chen, and Wei Wang.
\newblock Lion: Adversarial distillation of closed-source large language model.
\newblock \emph{CoRR}, abs/2305.12870, 2023.
\newblock \doi{10.48550/ARXIV.2305.12870}.
\newblock URL \url{https://doi.org/10.48550/arXiv.2305.12870}.

\bibitem[Jiayang et~al.(2024)Jiayang, Qiu, Chan, Liu, Song, and Zhang]{jiayang2024eventground}
Cheng Jiayang, Lin Qiu, Chunkit Chan, Xin Liu, Yangqiu Song, and Zheng Zhang.
\newblock Eventground: Narrative reasoning by grounding to eventuality-centric knowledge graphs, 2024.

\bibitem[Kosinski(2023)]{DBLP:journals/corr/abs-2302-02083}
Michal Kosinski.
\newblock Theory of mind may have spontaneously emerged in large language models.
\newblock \emph{CoRR}, abs/2302.02083, 2023.
\newblock \doi{10.48550/ARXIV.2302.02083}.
\newblock URL \url{https://doi.org/10.48550/arXiv.2302.02083}.

\bibitem[Li et~al.(2023{\natexlab{a}})Li, Chen, Luo, Kang, Zhang, Hu, Chan, and Song]{DBLP:journals/corr/abs-2310-10383}
Haoran Li, Yulin Chen, Jinglong Luo, Yan Kang, Xiaojin Zhang, Qi~Hu, Chunkit Chan, and Yangqiu Song.
\newblock Privacy in large language models: Attacks, defenses and future directions.
\newblock \emph{CoRR}, abs/2310.10383, 2023{\natexlab{a}}.
\newblock \doi{10.48550/ARXIV.2310.10383}.
\newblock URL \url{https://doi.org/10.48550/arXiv.2310.10383}.

\bibitem[Li et~al.(2023{\natexlab{b}})Li, Guo, Li, Fan, Hu, Liu, Chan, Yao, and Song]{DBLP:journals/corr/abs-2311-04044}
Haoran Li, Dadi Guo, Donghao Li, Wei Fan, Qi~Hu, Xin Liu, Chunkit Chan, Duanyi Yao, and Yangqiu Song.
\newblock P-bench: {A} multi-level privacy evaluation benchmark for language models.
\newblock \emph{CoRR}, abs/2311.04044, 2023{\natexlab{b}}.
\newblock \doi{10.48550/ARXIV.2311.04044}.
\newblock URL \url{https://doi.org/10.48550/arXiv.2311.04044}.

\bibitem[Li et~al.(2024)Li, Chen, Zheng, Hu, Chan, Liu, and Song]{DBLP:journals/corr/abs-2405-07667}
Haoran Li, Yulin Chen, Zihao Zheng, Qi~Hu, Chunkit Chan, Heshan Liu, and Yangqiu Song.
\newblock Backdoor removal for generative large language models.
\newblock \emph{CoRR}, abs/2405.07667, 2024.
\newblock \doi{10.48550/ARXIV.2405.07667}.
\newblock URL \url{https://doi.org/10.48550/arXiv.2405.07667}.

\bibitem[Li et~al.(2023{\natexlab{c}})Li, Chong, Stepputtis, Campbell, Hughes, Lewis, and Sycara]{li2023theory}
Huao Li, Yu~Quan Chong, Simon Stepputtis, Joseph Campbell, Dana Hughes, Michael Lewis, and Katia Sycara.
\newblock Theory of mind for multi-agent collaboration via large language models.
\newblock \emph{arXiv preprint arXiv:2310.10701}, 2023{\natexlab{c}}.

\bibitem[Lim et~al.(2020)Lim, Tio, and Ong]{DBLP:conf/cogsci/LimTO20}
Terence~X. Lim, Sidney Tio, and Desmond~C. Ong.
\newblock Improving multi-agent cooperation using theory of mind.
\newblock In Stephanie Denison, Michael Mack, Yang Xu, and Blair~C. Armstrong (eds.), \emph{Proceedings of the 42th Annual Meeting of the Cognitive Science Society - Developing a Mind: Learning in Humans, Animals, and Machines, CogSci 2020, virtual, July 29 - August 1, 2020}. cognitivesciencesociety.org, 2020.
\newblock URL \url{https://cogsci.mindmodeling.org/2020/papers/0195/index.html}.

\bibitem[Lu et~al.(2023)Lu, Zhao, Zhou, Li, et~al.]{lu2023danzero}
Yudong Lu, Youpeng Zhao, Wengang Zhou, Houqiang Li, et~al.
\newblock Danzero: Mastering guandan game with reinforcement learning.
\newblock In \emph{2023 IEEE Conference on Games (CoG)}, pp.\  1--8. IEEE, 2023.

\bibitem[Lukas et~al.(2023)Lukas, Salem, Sim, Tople, Wutschitz, and B{\'{e}}guelin]{DBLP:journals/corr/abs-2302-00539}
Nils Lukas, Ahmed Salem, Robert Sim, Shruti Tople, Lukas Wutschitz, and Santiago~Zanella B{\'{e}}guelin.
\newblock Analyzing leakage of personally identifiable information in language models.
\newblock \emph{CoRR}, abs/2302.00539, 2023.

\bibitem[Moghaddam \& Honey(2023)Moghaddam and Honey]{DBLP:journals/corr/abs-2304-11490}
Shima~Rahimi Moghaddam and Christopher~J. Honey.
\newblock Boosting theory-of-mind performance in large language models via prompting.
\newblock \emph{CoRR}, abs/2304.11490, 2023.
\newblock \doi{10.48550/ARXIV.2304.11490}.
\newblock URL \url{https://doi.org/10.48550/arXiv.2304.11490}.

\bibitem[Moravc{\'i}k et~al.(2017)Moravc{\'i}k, Schmid, Burch, Lis{\'y}, Morrill, Bard, Davis, Waugh, Johanson, and Bowling]{Moravck2017DeepStackEA}
Matej Moravc{\'i}k, Martin Schmid, Neil Burch, V.~Lis{\'y}, Dustin Morrill, Nolan Bard, Trevor Davis, K.~Waugh, Michael~Bradley Johanson, and Michael~H. Bowling.
\newblock Deepstack: Expert-level artificial intelligence in heads-up no-limit poker.
\newblock \emph{Science}, 356:\penalty0 508 -- 513, 2017.
\newblock URL \url{https://api.semanticscholar.org/CorpusID:1586260}.

\bibitem[Oguntola et~al.(2023)Oguntola, Campbell, Stepputtis, and Sycara]{DBLP:journals/corr/abs-2307-01158}
Ini Oguntola, Joseph Campbell, Simon Stepputtis, and Katia~P. Sycara.
\newblock Theory of mind as intrinsic motivation for multi-agent reinforcement learning.
\newblock \emph{CoRR}, abs/2307.01158, 2023.
\newblock \doi{10.48550/ARXIV.2307.01158}.
\newblock URL \url{https://doi.org/10.48550/arXiv.2307.01158}.

\bibitem[OpenAI(2023{\natexlab{a}})]{DBLP:journals/corr/abs-2303-08774}
OpenAI.
\newblock {GPT-4} technical report.
\newblock \emph{CoRR}, abs/2303.08774, 2023{\natexlab{a}}.
\newblock \doi{10.48550/ARXIV.2303.08774}.
\newblock URL \url{https://doi.org/10.48550/arXiv.2303.08774}.

\bibitem[OpenAI(2023{\natexlab{b}})]{GPT4-report}
OpenAI.
\newblock {GPT-4} technical report.
\newblock \emph{CoRR}, abs/2303.08774, 2023{\natexlab{b}}.
\newblock \doi{10.48550/ARXIV.2303.08774}.
\newblock URL \url{https://doi.org/10.48550/arXiv.2303.08774}.

\bibitem[OpenAI(2022)]{openai2022chatgpt}
TB~OpenAI.
\newblock Chatgpt: Optimizing language models for dialogue.
\newblock \emph{OpenAI}, 2022.

\bibitem[Ouyang et~al.(2022)Ouyang, Wu, Jiang, Almeida, Wainwright, Mishkin, Zhang, Agarwal, Slama, Ray, Schulman, Hilton, Kelton, Miller, Simens, Askell, Welinder, Christiano, Leike, and Lowe]{DBLP:conf/nips/Ouyang0JAWMZASR22}
Long Ouyang, Jeffrey Wu, Xu~Jiang, Diogo Almeida, Carroll~L. Wainwright, Pamela Mishkin, Chong Zhang, Sandhini Agarwal, Katarina Slama, Alex Ray, John Schulman, Jacob Hilton, Fraser Kelton, Luke Miller, Maddie Simens, Amanda Askell, Peter Welinder, Paul~F. Christiano, Jan Leike, and Ryan Lowe.
\newblock Training language models to follow instructions with human feedback.
\newblock In Sanmi Koyejo, S.~Mohamed, A.~Agarwal, Danielle Belgrave, K.~Cho, and A.~Oh (eds.), \emph{Advances in Neural Information Processing Systems 35: Annual Conference on Neural Information Processing Systems 2022, NeurIPS 2022, New Orleans, LA, USA, November 28 - December 9, 2022}, 2022.
\newblock URL \url{http://papers.nips.cc/paper\_files/paper/2022/hash/b1efde53be364a73914f58805a001731-Abstract-Conference.html}.

\bibitem[Park et~al.(2023)Park, O'Brien, Cai, Morris, Liang, and Bernstein]{DBLP:conf/uist/ParkOCMLB23}
Joon~Sung Park, Joseph~C. O'Brien, Carrie~Jun Cai, Meredith~Ringel Morris, Percy Liang, and Michael~S. Bernstein.
\newblock Generative agents: Interactive simulacra of human behavior.
\newblock In Sean Follmer, Jeff Han, J{\"{u}}rgen Steimle, and Nathalie~Henry Riche (eds.), \emph{Proceedings of the 36th Annual {ACM} Symposium on User Interface Software and Technology, {UIST} 2023, San Francisco, CA, USA, 29 October 2023- 1 November 2023}, pp.\  2:1--2:22. {ACM}, 2023.
\newblock \doi{10.1145/3586183.3606763}.
\newblock URL \url{https://doi.org/10.1145/3586183.3606763}.

\bibitem[Premack(1978)]{Premack1978DoesTC}
David Premack.
\newblock Does the chimpanzee have a theory of mind?
\newblock \emph{Behavioral and Brain Sciences}, 1:\penalty0 515 -- 526, 1978.
\newblock URL \url{https://api.semanticscholar.org/CorpusID:141321709}.

\bibitem[Raman et~al.(2022)Raman, Cohen, Rosen, Idrees, Paulius, and Tellex]{DBLP:journals/corr/abs-2211-09935}
Shreyas~Sundara Raman, Vanya Cohen, Eric Rosen, Ifrah Idrees, David Paulius, and Stefanie Tellex.
\newblock Planning with large language models via corrective re-prompting.
\newblock \emph{CoRR}, abs/2211.09935, 2022.
\newblock \doi{10.48550/ARXIV.2211.09935}.
\newblock URL \url{https://doi.org/10.48550/arXiv.2211.09935}.

\bibitem[Riedl et~al.(2021)Riedl, Kim, Gupta, Malone, and Woolley]{Riedl2021QuantifyingCI}
Christoph Riedl, Young~Ji Kim, Pranav Gupta, Thomas~W. Malone, and Anita~Williams Woolley.
\newblock Quantifying collective intelligence in human groups.
\newblock \emph{Proceedings of the National Academy of Sciences}, 118, 2021.
\newblock URL \url{https://api.semanticscholar.org/CorpusID:234769649}.

\bibitem[Sclar et~al.(2023)Sclar, Kumar, West, Suhr, Choi, and Tsvetkov]{DBLP:conf/acl/SclarKWS0T23}
Melanie Sclar, Sachin Kumar, Peter West, Alane Suhr, Yejin Choi, and Yulia Tsvetkov.
\newblock Minding language models' (lack of) theory of mind: {A} plug-and-play multi-character belief tracker.
\newblock In Anna Rogers, Jordan~L. Boyd{-}Graber, and Naoaki Okazaki (eds.), \emph{Proceedings of the 61st Annual Meeting of the Association for Computational Linguistics (Volume 1: Long Papers), {ACL} 2023, Toronto, Canada, July 9-14, 2023}, pp.\  13960--13980. Association for Computational Linguistics, 2023.
\newblock \doi{10.18653/V1/2023.ACL-LONG.780}.
\newblock URL \url{https://doi.org/10.18653/v1/2023.acl-long.780}.

\bibitem[Susnjak(2022)]{DBLP:journals/corr/abs-2212-09292}
Teo Susnjak.
\newblock Chatgpt: The end of online exam integrity?
\newblock \emph{CoRR}, abs/2212.09292, 2022.

\bibitem[Taori et~al.(2023)Taori, Gulrajani, Zhang, Dubois, Li, Guestrin, Liang, and Hashimoto]{alpaca}
Rohan Taori, Ishaan Gulrajani, Tianyi Zhang, Yann Dubois, Xuechen Li, Carlos Guestrin, Percy Liang, and Tatsunori~B. Hashimoto.
\newblock Stanford alpaca: An instruction-following llama model.
\newblock \url{https://github.com/tatsu-lab/stanford_alpaca}, 2023.

\bibitem[Touvron et~al.(2023)Touvron, Martin, Stone, Albert, Almahairi, Babaei, Bashlykov, Batra, Bhargava, Bhosale, Bikel, Blecher, Canton{-}Ferrer, Chen, Cucurull, Esiobu, Fernandes, Fu, Fu, Fuller, Gao, Goswami, Goyal, Hartshorn, Hosseini, Hou, Inan, Kardas, Kerkez, Khabsa, Kloumann, Korenev, Koura, Lachaux, Lavril, Lee, Liskovich, Lu, Mao, Martinet, Mihaylov, Mishra, Molybog, Nie, Poulton, Reizenstein, Rungta, Saladi, Schelten, Silva, Smith, Subramanian, Tan, Tang, Taylor, Williams, Kuan, Xu, Yan, Zarov, Zhang, Fan, Kambadur, Narang, Rodriguez, Stojnic, Edunov, and Scialom]{DBLP:journals/corr/abs-2307-09288}
Hugo Touvron, Louis Martin, Kevin Stone, Peter Albert, Amjad Almahairi, Yasmine Babaei, Nikolay Bashlykov, Soumya Batra, Prajjwal Bhargava, Shruti Bhosale, Dan Bikel, Lukas Blecher, Cristian Canton{-}Ferrer, Moya Chen, Guillem Cucurull, David Esiobu, Jude Fernandes, Jeremy Fu, Wenyin Fu, Brian Fuller, Cynthia Gao, Vedanuj Goswami, Naman Goyal, Anthony Hartshorn, Saghar Hosseini, Rui Hou, Hakan Inan, Marcin Kardas, Viktor Kerkez, Madian Khabsa, Isabel Kloumann, Artem Korenev, Punit~Singh Koura, Marie{-}Anne Lachaux, Thibaut Lavril, Jenya Lee, Diana Liskovich, Yinghai Lu, Yuning Mao, Xavier Martinet, Todor Mihaylov, Pushkar Mishra, Igor Molybog, Yixin Nie, Andrew Poulton, Jeremy Reizenstein, Rashi Rungta, Kalyan Saladi, Alan Schelten, Ruan Silva, Eric~Michael Smith, Ranjan Subramanian, Xiaoqing~Ellen Tan, Binh Tang, Ross Taylor, Adina Williams, Jian~Xiang Kuan, Puxin Xu, Zheng Yan, Iliyan Zarov, Yuchen Zhang, Angela Fan, Melanie Kambadur, Sharan Narang, Aur{\'{e}}lien Rodriguez, Robert Stojnic, Sergey Edunov,
  and Thomas Scialom.
\newblock Llama 2: Open foundation and fine-tuned chat models.
\newblock \emph{CoRR}, abs/2307.09288, 2023.
\newblock \doi{10.48550/ARXIV.2307.09288}.
\newblock URL \url{https://doi.org/10.48550/arXiv.2307.09288}.

\bibitem[Ullman(2023)]{ullman2023large}
Tomer Ullman.
\newblock Large language models fail on trivial alterations to theory-of-mind tasks.
\newblock \emph{arXiv preprint arXiv:2302.08399}, 2023.

\bibitem[Wang et~al.(2023)Wang, Xie, Jiang, Mandlekar, Xiao, Zhu, Fan, and Anandkumar]{DBLP:journals/corr/abs-2305-16291}
Guanzhi Wang, Yuqi Xie, Yunfan Jiang, Ajay Mandlekar, Chaowei Xiao, Yuke Zhu, Linxi Fan, and Anima Anandkumar.
\newblock Voyager: An open-ended embodied agent with large language models.
\newblock \emph{CoRR}, abs/2305.16291, 2023.
\newblock \doi{10.48550/ARXIV.2305.16291}.
\newblock URL \url{https://doi.org/10.48550/arXiv.2305.16291}.

\bibitem[Wang et~al.(2022)Wang, Wu, Evans, Parkes, Tenenbaum, and Kleiman{-}Weiner]{DBLP:books/ox/22/WangWEPTK22}
Rose~E. Wang, Sarah~A. Wu, James~A. Evans, David~C. Parkes, Joshua~B. Tenenbaum, and Max Kleiman{-}Weiner.
\newblock Too many cooks: Bayesian inference for coordinating multi-agent collaboration.
\newblock In Stephen~H. Muggleton and Nicholas Chater (eds.), \emph{Human-Like Machine Intelligence}, pp.\  152--170. Oxford University Press, 2022.
\newblock \doi{10.1093/OSO/9780198862536.003.0008}.
\newblock URL \url{https://doi.org/10.1093/oso/9780198862536.003.0008}.

\bibitem[Wang et~al.(2024)Wang, Fang, Li, Shi, Ding, Xu, Wang, Bai, Liu, Cheng, Chan, and Song]{DBLP:journals/corr/abs-2401-07286}
Weiqi Wang, Tianqing Fang, Chunyang Li, Haochen Shi, Wenxuan Ding, Baixuan Xu, Zhaowei Wang, Jiaxin Bai, Xin Liu, Jiayang Cheng, Chunkit Chan, and Yangqiu Song.
\newblock {CANDLE:} iterative conceptualization and instantiation distillation from large language models for commonsense reasoning.
\newblock \emph{CoRR}, abs/2401.07286, 2024.
\newblock \doi{10.48550/ARXIV.2401.07286}.
\newblock URL \url{https://doi.org/10.48550/arXiv.2401.07286}.

\bibitem[Wei et~al.(2022)Wei, Wang, Schuurmans, Bosma, Ichter, Xia, Chi, Le, and Zhou]{DBLP:conf/nips/Wei0SBIXCLZ22}
Jason Wei, Xuezhi Wang, Dale Schuurmans, Maarten Bosma, Brian Ichter, Fei Xia, Ed~H. Chi, Quoc~V. Le, and Denny Zhou.
\newblock Chain-of-thought prompting elicits reasoning in large language models.
\newblock In Sanmi Koyejo, S.~Mohamed, A.~Agarwal, Danielle Belgrave, K.~Cho, and A.~Oh (eds.), \emph{Advances in Neural Information Processing Systems 35: Annual Conference on Neural Information Processing Systems 2022, NeurIPS 2022, New Orleans, LA, USA, November 28 - December 9, 2022}, 2022.
\newblock URL \url{http://papers.nips.cc/paper\_files/paper/2022/hash/9d5609613524ecf4f15af0f7b31abca4-Abstract-Conference.html}.

\bibitem[Wu et~al.(2023)Wu, Prabhumoye, Min, Bisk, Salakhutdinov, Azaria, Mitchell, and Li]{DBLP:journals/corr/abs-2305-15486}
Yue Wu, Shrimai Prabhumoye, So~Yeon Min, Yonatan Bisk, Ruslan Salakhutdinov, Amos Azaria, Tom~M. Mitchell, and Yuanzhi Li.
\newblock {SPRING:} {GPT-4} out-performs {RL} algorithms by studying papers and reasoning.
\newblock \emph{CoRR}, abs/2305.15486, 2023.
\newblock \doi{10.48550/ARXIV.2305.15486}.
\newblock URL \url{https://doi.org/10.48550/arXiv.2305.15486}.

\bibitem[Yang et~al.(2023)Yang, Xiao, Wang, Zhang, Bian, Yin, Lv, Pan, Wang, Yan, Yang, Deng, Wang, Liu, Ai, Dong, Zhao, Xu, Sun, Zhang, Liu, Ji, Xie, Dai, Fang, Su, Song, Liu, Ru, Ma, Wang, Liu, Lin, Nie, Guo, Sun, Zhang, Li, Li, Cheng, Chen, Zeng, Wang, Chen, Men, Yu, Pan, Shen, Wang, Li, Jiang, Gao, Zhang, Zhou, and Wu]{baichuan2}
Aiyuan Yang, Bin Xiao, Bingning Wang, Borong Zhang, Ce~Bian, Chao Yin, Chenxu Lv, Da~Pan, Dian Wang, Dong Yan, Fan Yang, Fei Deng, Feng Wang, Feng Liu, Guangwei Ai, Guosheng Dong, Haizhou Zhao, Hang Xu, Haoze Sun, Hongda Zhang, Hui Liu, Jiaming Ji, Jian Xie, Juntao Dai, Kun Fang, Lei Su, Liang Song, Lifeng Liu, Liyun Ru, Luyao Ma, Mang Wang, Mickel Liu, MingAn Lin, Nuolan Nie, Peidong Guo, Ruiyang Sun, Tao Zhang, Tianpeng Li, Tianyu Li, Wei Cheng, Weipeng Chen, Xiangrong Zeng, Xiaochuan Wang, Xiaoxi Chen, Xin Men, Xin Yu, Xuehai Pan, Yanjun Shen, Yiding Wang, Yiyu Li, Youxin Jiang, Yuchen Gao, Yupeng Zhang, Zenan Zhou, and Zhiying Wu.
\newblock Baichuan 2: Open large-scale language models.
\newblock \emph{CoRR}, abs/2309.10305, 2023.
\newblock \doi{10.48550/ARXIV.2309.10305}.
\newblock URL \url{https://doi.org/10.48550/arXiv.2309.10305}.

\bibitem[Yao et~al.(2023)Yao, Yu, Zhao, Shafran, Griffiths, Cao, and Narasimhan]{DBLP:conf/nips/YaoYZS00N23}
Shunyu Yao, Dian Yu, Jeffrey Zhao, Izhak Shafran, Tom Griffiths, Yuan Cao, and Karthik Narasimhan.
\newblock Tree of thoughts: Deliberate problem solving with large language models.
\newblock In Alice Oh, Tristan Naumann, Amir Globerson, Kate Saenko, Moritz Hardt, and Sergey Levine (eds.), \emph{Advances in Neural Information Processing Systems 36: Annual Conference on Neural Information Processing Systems 2023, NeurIPS 2023, New Orleans, LA, USA, December 10 - 16, 2023}, 2023.
\newblock URL \url{http://papers.nips.cc/paper\_files/paper/2023/hash/271db9922b8d1f4dd7aaef84ed5ac703-Abstract-Conference.html}.

\bibitem[Yuan et~al.(2021)Yuan, Fu, Zhou, Yang, and Zhu]{DBLP:journals/corr/abs-2110-00121}
Luyao Yuan, Zipeng Fu, Linqi Zhou, Kexin Yang, and Song{-}Chun Zhu.
\newblock Emergence of theory of mind collaboration in multiagent systems.
\newblock \emph{CoRR}, abs/2110.00121, 2021.
\newblock URL \url{https://arxiv.org/abs/2110.00121}.

\bibitem[Zeng et~al.(2023)Zeng, Liu, Du, Wang, Lai, Ding, Yang, Xu, Zheng, Xia, Tam, Ma, Xue, Zhai, Chen, Liu, Zhang, Dong, and Tang]{chatglm}
Aohan Zeng, Xiao Liu, Zhengxiao Du, Zihan Wang, Hanyu Lai, Ming Ding, Zhuoyi Yang, Yifan Xu, Wendi Zheng, Xiao Xia, Weng~Lam Tam, Zixuan Ma, Yufei Xue, Jidong Zhai, Wenguang Chen, Zhiyuan Liu, Peng Zhang, Yuxiao Dong, and Jie Tang.
\newblock {GLM-130B:} an open bilingual pre-trained model.
\newblock In \emph{The Eleventh International Conference on Learning Representations, {ICLR} 2023, Kigali, Rwanda, May 1-5, 2023}. OpenReview.net, 2023.
\newblock URL \url{https://openreview.net/pdf?id=-Aw0rrrPUF}.

\bibitem[Zha et~al.(2019)Zha, Lai, Cao, Huang, Wei, Guo, and Hu]{RLCard}
Daochen Zha, Kwei{-}Herng Lai, Yuanpu Cao, Songyi Huang, Ruzhe Wei, Junyu Guo, and Xia Hu.
\newblock Rlcard: {A} toolkit for reinforcement learning in card games.
\newblock \emph{CoRR}, abs/1910.04376, 2019.
\newblock URL \url{http://arxiv.org/abs/1910.04376}.

\bibitem[Zhao et~al.(2023)Zhao, Lu, Zhao, Zhou, and Li]{zhao2023danzero+}
Youpeng Zhao, Yudong Lu, Jian Zhao, Wengang Zhou, and Houqiang Li.
\newblock Danzero+: Dominating the guandan game through reinforcement learning.
\newblock \emph{arXiv preprint arXiv:2312.02561}, 2023.

\end{thebibliography}
\bibliographystyle{iclr2024_conference}

\appendix
\appendix
\section{Guan Dan Rules}\label{app:a}
Guandan is a popular team-based, competitive card game in China. The gameplay is somewhat similar to Doudizhu:

\subsection{English version: }

1. Game Setup:
\begin{itemize}
\item Number of players: 4
\item Number of cards used: Two decks of 54 cards each, totaling 108 cards
\item Number of cards per player: 27
\item Team division: Players sitting opposite each other form a team, while neighboring players are on the opposing team
\end{itemize}

2. Card rankings and suits:
\begin{itemize}
\item Card ranks from low to high: 2, 3, 4, 5, 6, 7, 8, 9, 10, J, Q, K, A, Black Joker, Red Joker
\item Suits: Hearts ($\heartsuit$), Spades ($\spadesuit$), Clubs ($\clubsuit$), Diamonds ($\diamondsuit$)
\item Special card: The Hearts card of the current round's rank can be used as any card point (except Jokers) to form combinations, known as ``Feng Ren Pei'' (Matching Card).
\end{itemize}

3. Playing order:
\begin{itemize}
\item In one round, the playing order is: Player 0 $\rightarrow$ Player 1 $\rightarrow$ Player 2 $\rightarrow$ Player 3 $\rightarrow$ Player 0 ...
\item Players must play a higher-ranking combination than the previous one, until all other players pass, which concludes the round
\end{itemize}

4. Card Combinations:
Divided into two categories:
\begin{enumerate}
\item Category One:
\begin{itemize}
\item Single: Any single card
\item Pair: Any two cards of the same rank, including a pair of Jokers or Big Jokers
\item Wood Board: Three pairs of adjacent ranks, such as 223344
\item Trips: Three cards of the same rank
\item Steel Board: Two sets of trips with the same rank, such as 333444
\item Trips with a Pair: Trips plus a pair, such as 55522
\item Straight: Five cards of consecutive ranks, such as 23456
\item Different combinations in this category cannot beat each other
\end{itemize}
\item Category Two:
\begin{itemize}
    \item Bomb: Four or more cards of the same rank
    \item Flush: A straight flush consisting of five cards of the same suit
    \item Heavenly Bomb: Two Jokers and two Big Jokers
\end{itemize}

Card type Rankings:
\begin{itemize}
    \item In the first category, the combination with higher card points wins. For Trips with a Pair, only the rank of the Trips is compared. Ace is the lowest card (1) in Three Pairs, Two Trips, Straight, and Flush.
    \item Category Two can beat any combination in Category One. The ranking within Category Two is as follows:
    \begin{itemize}
        \item A Bomb with more cards beats one with fewer. If the number of cards is the same, the one with higher ranks wins.
        \item For Flushes, the one with higher card points wins. A Flush can beat a Bomb of five cards or less, but not a Bomb of six or more.
        \item The Heavenly Bomb is the highest and can beat all other combinations.
    \end{itemize}
\end{itemize}
\end{enumerate}

5. Game Terminology:
\begin{itemize}
\item One Deal: The process from dealing cards to all players playing their hands
\item Finishing order terms: First Place (first to finish), Second Place, Third Place, Last Place (last to finish)
\item Double Last: When both players on a team finish in Third and Last places, which is disadvantageous for the next deal
\end{itemize}

6. Scoring Rules (per deal): \label{app:a1}
\begin{itemize}
\item Double First (one team's players finish in First and Second places): That team +4 points, the other team -4 points
\item One team's players finish in First and Third places: That team +2 points, the other team -2 points
\item One team's players finish in First and Last places: That team +1 point, the other team -1 point
\item Note: Teammates always have the same score, reflecting the importance of teamwork
\end{itemize}

7. Winning Condition:
After several rounds, the team with the highest total score wins. Getting First Place is crucial for scoring and for winning the game.

\subsection{Chinese version: } \label{app:a2}

\begin{mybox1}
\textbf{Sample prompt to LLM: }
\begin{CJK}{UTF8}{gkai}
\subsection*{1. 游戏设置:}
\begin{itemize}
\item 游戏人数: 4人
\item 使用牌数: 两副54张牌,共108张
\item 每人手牌数: 27张
\item 队伍划分: 相对的两人为一方,上家和下家为对方
\end{itemize}

2. 牌点大小和花色:
\begin{itemize}
\item 牌点由小到大: 2, 3, 4, 5, 6, 7, 8, 9, 10, J, Q, K, A, 小王, 大王
\item 花色: 红桃 ($\heartsuit$), 黑桃 ($\spadesuit$), 梅花 ($\clubsuit$), 方块 ($\diamondsuit$)
\item 特殊牌: 红心级牌可作为任意牌点的花色牌 (大小王除外) 搭配使用, 称为 "逢人配"。每回合的逢人配由当局级牌的红心牌担任。
\end{itemize}

\subsection*{3. 出牌顺序:}
\begin{itemize}
\item 一圈牌中, 出牌顺序为玩家索引0 $\rightarrow$ 玩家索引1 $\rightarrow$ 玩家索引2 $\rightarrow$ 玩家索引3 $\rightarrow$ 玩家索引0 ...
\item 以更大的牌型压制前手牌, 直至其他玩家不出为止, 称为一圈牌
\end{itemize}

4. 牌型:\
分为两大类:
\begin{itemize}
\item 第一类:
\begin{itemize}
\item 单张:任意一张单牌
\item 对子:任意两张点数相同的牌,包括对大王或对小王
\item 木板:三对点数相邻的对子,如223344
\item 三同张:三张点数相同的牌
\item 钢板:两套点数相同的三同张,如333444
\item 三带二:三同张加一个对子,如55522
\item 顺子:五张点数相邻的牌,如23456
\item 以上不同牌型之间不能相互压制
\end{itemize}

\item 第二类:
\begin{itemize}
\item 炸弹:四张或四张以上点数相同的牌
\item 同花顺:五张花色相同的顺子
\item 天王炸:大小王各两张
\item 牌型大小:
\item 第一类牌型中,以点数大者为大。三带二只比较三同张的大小。2在三连对、二连三、顺子、同花顺中视为1点,最小。
\item 炸弹张数多者为大，若张数相同，则点数大者为大
\item 同花顺以点数大者为大，可压制五张及以下的炸弹，但不能压制六张及以上的炸弹
\item 天王炸为最大，可压制所有牌型
\end{itemize}
\end{itemize}
\end{CJK}
\end{mybox1}

\begin{mybox1}
\begin{CJK}{UTF8}{gkai}

5. 游戏术语:
\begin{itemize}
\item 一副牌:从开始发牌到所有玩家出完手中牌的过程
\item 出完牌顺序称谓:上游(第一个出完)、二游、三游、下游(最后一个出完)
\item 双下游:一方两人分别为三游和下游,对下一副牌的局势较为不利
\end{itemize}

6. 计分规则(每副牌):
\begin{itemize}
\item 双上(一方两人分别为上游和二游):该方+4分,另一方-4分
\item 一方两人分别为上游和三游:该方+2分,另一方-2分
\item 一方两人分别为上游和下游:该方+1分,另一方-1分
\item 注:同一方两人得分总是相同,体现团队合作精神
\end{itemize}

7. 获胜条件:\
经过若干局后,总分较高的一方获胜。由于上游得分至关重要,因此争取上游是取胜的关键。

\end{CJK}
\end{mybox1}

\section{Prompt template Samples} \label{app:b}
\subsection{Sample Prompts for First-Order Behaviour Pattern Analysis (Incomplete): } \label{app:b1}

\begin{tcolorbox}[colback=gray!20]
\textit{Understand the rules of the game, the cards you have, your observations, and a summary of the current game history. Please infer the beliefs of all players on the field and the advantages and disadvantages of the cards they hold, for example, they may lack a card type. }

\begin{CJK}{UTF8}{gkai}
了解游戏规则、你拥有的牌、你的观察结果和总结当前游戏历史记录。请推断出场上所有玩家的信念以及他们所持有的手牌的优势和劣势，例如，他们可能缺乏卡的种类。
\end{CJK}
\end{tcolorbox}

\subsection{Sample Prompts for Planning and Evaluator (Incomplete):} \label{app:b2}

\begin{tcolorbox}[colback=gray!20]
\textit{Develop a rational strategy: Considering the available actions, formulate a plan for each action to progressively achieve victory. From the adversary's viewpoint, deduce the categories of cards in their hand and their team's collaborative tactics based on their behavioral patterns, predict their likely actions, and infer teammates' beliefs and hand cards to construct a cooperative strategy. }

\textit{Estimate the expected gain for each plan: Estimate the final points your team may gain by choosing this plan.}

\begin{CJK}{UTF8}{gkai}
策略：考虑可采取的行动，为每项行动制定计划，逐步取得胜利。从对手的角度，推断出对手手中牌的类别以及他们团队的协作策略（基于他们以往的行为模式），预测他们可能的行动，并推断队友的信念和手牌以构建合作策略。

估计每个计划的预期收益：估计您的团队选择这个计划可能获得的最终分数。
\end{CJK}
\end{tcolorbox}

\subsection{Sample Prompts for Second-Order Behaviour Pattern Analysis (Incomplete):} \label{app:b3}
\begin{tcolorbox}[colback=gray!20]
\textit{From my perspective, please analyze the impact and belief of each plan on the subsequent trend of the game, taking into account the possible hand conditions of the two other opponents and allies.
From the opponent's point of view (being able to observe only my actions, not my cards), please analyze their potential beliefs about my hand and our cooperative strategy.}

\begin{CJK}{UTF8}{gkai}
从我的角度来看，请分析每个计划对后续比赛走势的影响和信念，同时考虑到另外一位盟友和两位对手可能的手牌情况。
从对手的角度（只能观察我的行动，而不是我的牌），请分析他们对我的手牌和我们的合作策略的潜在信念。
\end{CJK}
\end{tcolorbox}

\section{Output Samples} \label{app:c}
\subsection{Convert game state to readable text}\label{app:c1}
\begin{mybox}
\textbf{Readable Current State Observation: }
\par
\textit{1. My current hand cards are: ['Diamond 2', 'Diamond 4', 'Club 3', 'Club 4', 'Club A', 'Club A', 'Heart K', 'Heart A', 'Spade 2', 'Spade 3', 'Spade 4', 'Spade 8', 'Spade J'], and the remaining number of hand cards is 13. The current Joker card is Heart J. The Joker is a special card type that can act as any point card of any suit, helping to form larger card combinations.}

\textit{2. My teammate (with index number 2) has a total of 8 hand cards.}

\textit{3. If I play a card, the opponent (with index number 0) who is ready to follow me has 2 hand cards remaining.}

\textit{4. My previous opponent (with index number 3) has 13 hand cards.}

\textit{5. I currently do not have the right to play a card.}

\textit{6. The last player to play a card was the opponent (player index 2), and the card type he played was a pair, specifically ['Diamond 9', 'Heart 9'], and he has 2 hand cards remaining. }

\medskip

\begin{CJK}{UTF8}{gkai}
1. 我目前拥有的手牌为：['方片 2', '方片 4', '梅花 3', '梅花 4', '梅花 A', '梅花 A', '红心 K', '红心 A', '黑桃 2', '黑桃 3', '黑桃 4', '黑桃 8', '黑桃 J']，剩余手牌数量为13，当前逢人配为红心 J。逢人配是一种特殊的牌型，可以充当任意牌点的花色牌，帮助组合出更大的牌型。

2. 我的队友（索引号为2）所有的手牌数量为8。

3. 如果我打出牌之后，准备要接我的对手（索引号为0），他所拥有的手牌数量为2。

4. 我的上家对手（索引号为3）所拥有的手牌数量为13。

5. 我当前没有出牌权。

6. 上次出牌的玩家为对手（玩家索引2），他打出的牌型为对牌，具体牌为['方片 9', '红心 9']，他所剩余的手牌数量为2。

\end{CJK}
\end{mybox}

\subsection{History record to readable text}\label{app:c2}

\begin{mybox}
\textbf{Readable History record: }

\par

\textit{1. Player 2 (teammate) played a single card with the value ['Joker'].}

\textit{2. Player 3 (previous opponent) passed.}

\textit{3. Player 1 (next opponent) passed.}

\textit{4. Player 2 (teammate) played a pair with the values ['Diamond 4', 'Heart 4'].}

\textit{5. Player 3 (previous opponent) passed.}

\textit{6. Player 1 (next opponent) played a pair with the values ['Club 8', 'Heart 8'].}

\textit{7. Player 2 (teammate) played a pair with the values ['Diamond 9', 'Heart 9'].}

\textit{8. Player 3 (previous opponent) passed.}

\medskip

\begin{CJK}{UTF8}{gkai}
1. 2号玩家(队友)打出了牌值为 ['小王'] 的单牌。

2. 3号玩家(上家对手)过牌。

3. 1号玩家(下家对手)过牌。

4. 2号玩家(队友)打出了牌值为 ['方片 4', '红心 4'] 的对牌。

5. 3号玩家(上家对手)过牌。

6. 1号玩家(下家对手)打出了牌值为 ['梅花 8', '红心 8'] 的对牌。

7. 2号玩家(队友)打出了牌值为 ['方片 9', '红心 9'] 的对牌。

8. 3号玩家(上家对手)过牌。

\end{CJK}
\end{mybox}

\subsection{Planning module, formulates plans for each valid action}

\begin{mybox}
\textbf{Planning: }
\par
\textit{1. Convert the legal action list dictionary into readable text:
Legal action index 0: The card combination type is a pair, and the hand cards forming the combination are Club A, Club A.
Legal action index 1: The card combination type is pass, and there are no hand cards forming the combination.}

\medskip

\textit{2. Based on my beliefs, develop a set of subsequent plans:
Plan 0: If I play action index 0, it will suppress my teammate and cause them to lose the right to play. The core of my plan is to assist my teammate in achieving the first place.}

\textit{Plan 1: If I play action index 1, I can wait for better card combinations before playing, preserving more chips. The core of my plan is to assist my teammate and keep the pair of A's to prevent my teammate from being unable to follow suit later.}

\medskip

\textit{3. Analyze the impact of each plan on the subsequent game trends and beliefs:
If I adopt Plan 0, it will suppress my teammate, lose stronger card combinations, cause my teammate to lose the right to play, and put our team at a disadvantage. After playing this set of cards, the cards I may have left could be Club 2, Club 3, Club 4, Heart K, Heart A, Spade 2, Spade 3, Spade 8, and Spade J, which may not be easy to play later.
If I adopt Plan 1, it is a conservative approach to assist my teammates in finishing their hand, preserving the opportunity to suppress the opponents. After playing this set of cards, the number of cards I have left remains unchanged, and I will need to make further plans based on the situation.}

\medskip

\textit{4. Estimate the expected gain for each plan:
After adopting Plan 0, my teammate lost the right to play, and I also played the strong pair of A's. I expect this will cause our team to be at a disadvantage and unable to achieve first place. The total points our team can obtain may be lower than 1 point.}

\textit{After adopting Plan 1, my ranking when I finish all my cards could be 2nd, 3rd, or 4th place, as the assistive approach may miss the opportunity for first place. My teammate's expected ranking, when they finish their cards, could be first place, so we can obtain more points. The total points our team can obtain may be above 1 point.}

\medskip

\textit{5. Plan selection: Based on the expected gain ranking, choose the plan with the highest expected benefit.
According to the analysis, Plan 1 has the highest expected benefit because it can assist my teammate in passing, while preserving stronger card combinations, with the possibility of achieving first place. Therefore, Plan 1 is selected.}

\medskip
\end{mybox}
\begin{mybox}
\begin{CJK}{UTF8}{gkai}

1. 将合法操作列表字典转为可读文本：
合法动作索引0：牌型为对牌，组成牌型的手牌为梅花 A, 梅花 A。
合法动作索引1：牌型为过牌，组成牌型的手牌为无。

\medskip

2. 基于您的信念，制定一套后续方案：
方案0：我打出动作索引0的话，将会压制队友，导致队友失去出牌权。我的方案核心是辅助队友取得头游。
方案1：我打出动作索引1的话，可以等待更好的牌型再出牌，保留更多筹码。我的方案核心是辅助队友，保留对A防止后续队友接不住牌。

\medskip

3. 分析每套方案对比赛后续走势影响以及信念：
如果我采用方案0，会压制队友，失去更强大的牌型，让队友失去出牌权，导致队伍陷入劣势。打完这套牌之后我所剩下的牌可能还有梅花 2, 梅花 3, 梅花 4, 红心 K, 红心 A, 黑桃 2, 黑桃 3, 黑桃 8, 黑桃 J，这些牌后续不太好走掉。
如果我采用方案1，保守打法，辅助队友打完手牌，保留压制对手的机会。打完这套牌之后我所剩下的牌数量不变，后续需要根据情况再做打算。

\medskip

4. 估算每套方案的预期增益：
采用方案0后，队友失去出牌权，且我也打出了强大的对A。我预期这样会导致队伍陷入劣势，没办法成为头游。队伍可以获得的总分可能是低于1分。
采用方案1后，我走完所有手牌的名次可能是2,3,4游，因为辅助打法可能错失头游的机会。我队友预期走完牌的名次可能是头游，因此我们可以取得更多分数。队伍可以获得的总分可能是1分以上。

\medskip

5. 方案选择：根据预期增益排名，选择预期获益最高的方案。
根据分析，方案1的预期获益最高，因为可以辅助队友过牌，同时保留更强大的牌型，有可能取得头游，因此选择方案1。

\end{CJK}
\end{mybox}

\newpage
\subsection{Acquire second-order ToM knowledge} \label{app:c4}
\begin{mybox}
\textbf{Second-order ToM: }
\par
\textit{Based on my current hand ['Diamond 2', 'Diamond 4', 'Club 3', 'Club 4', 'Club A', 'Club A', 'Heart K', 'Heart A', 'Spade 2', 'Spade 3', 'Spade 4', 'Spade 8', 'Spade J'], I have a pair of Club A, a pair of 3, a pair of 4, pairs, a single card, and a wild card (Heart J). My hand lacks stronger card combinations such as straights, three-of-a-kind, full houses, bombs, etc., but the wild card can help me form larger card combinations.}

\medskip
\textit{Belief about the hand of player index 2 (teammate):
Based on the historical play records, my teammate played pairs and single cards in the previous rounds, with a hand size of 8. They may have some pairs and single cards, but the specific card types are difficult to determine. They may lack some stronger card combinations and need our cooperation to suppress the opponents, helping them be the first to play all their cards.}

\medskip

\textit{Player index 2's (teammate) potential belief of my current hand: My teammate may think that I have some pairs and single cards in my hand, but the specific card types are difficult to determine. In the previous rounds, after they played some smaller pairs, I did not follow their cards. They may understand that I lack decisive winning cards and are cooperating with them to let them become the first to finish.}

\medskip

\textit{Belief about our team's cooperation plan: I should currently assist my teammate in being the first player to finish all their cards, cooperating with each other and choosing a playing strategy based on the situation on the table and the opponents' play, striving to achieve first or second place to obtain more points.}

\medskip

\textit{Belief about the cards of player index 0 (enemy):
Based on the historical play records, enemy 1 did not follow my teammate's pair of 4 in the previous rounds, with a hand size of 2. They may have fewer cards, and it can be determined that they do not have pairs in their hand. They have some stronger single cards, and I need to choose appropriate card combinations to suppress them when playing.}

\medskip

\textit{Belief about the cards of player index 3 (enemy):
Based on the historical play records, enemy 2 passed in the previous rounds with a hand size of 13. They may have more cards, but the specific card types are difficult to determine. They may have some stronger card combinations and are trying their best to assist their teammate in becoming the first to finish. Therefore, they may play any strong cards in their hand to suppress us and then use single cards to help their teammate win.}

\medskip

\textit{Player index 0 and 3's (enemies) potential belief of my current hand: The enemies may think that I have some pairs and single cards in my hand, but the specific card types are difficult to determine. They may believe that I lack some stronger card combinations but may also be wary of the wild card in my hand, requiring me to choose carefully when playing.}

\medskip

\begin{CJK}{UTF8}{gkai}

根据我当前手牌['方片 2', '方片 4', '梅花 3', '梅花 4', '梅花 A', '梅花 A', '红心 K', '红心 A', '黑桃 2', '黑桃 3', '黑桃 4', '黑桃 8', '黑桃 J']，我拥有一对梅花 A，一对 3，一对 4，对子，一张单牌，一张逢人配（红心 J）。我的手牌中缺少顺子、三同张、钢板、炸弹等比较强大的牌型，但有逢人配可以帮助我组合出更大的牌型。

\medskip

对玩家索引 2（队友）的手牌的信念：
根据历史出牌记录，队友在前几轮中打出了对牌和单牌，手牌数量为8，可能拥有一些对子和单牌，但具体牌型难以确定。他可能缺少一些比较强大的牌型，需要我们合作来压制对手，配合他第一个打完所有手牌。

玩家索引为 2（队友）当前对我的手牌的潜在看法：
队友可能认为我手牌中拥有一些对子和单牌，但具体牌型难以确定。前几回合在他打了一些比较小对子之后，我没有接他的牌，他可能理解到我缺乏决定性获胜的手牌，正在配合他，让他成为头游。

\medskip

对我们团队配合计划的信念：
我目前应该辅助队友让他尽力成为第一个走完所有手牌的玩家，互相配合，根据场上局势和对手出牌情况来选择出牌策略，争取取得头游或者二游，以获取更多的分数。

\medskip
\end{CJK}
\end{mybox}

\begin{mybox}
\begin{CJK}{UTF8}{gkai}
对玩家索引 0（敌人）的牌的信念：
根据历史出牌记录，敌人1在前几轮中没有接住队友打出的对4 ，手牌数量为2，可能手牌较少，可以确定他手牌中不存在对子。他拥有一些比较强大的单牌，需要我在出牌时选择合适的牌型来压制他。

\medskip

对玩家索引 3（敌人）的牌的信念：
根据历史出牌记录，敌人2在前几轮中过牌，手牌数量为13，可能手牌较多，但具体牌型难以确定。他可能拥有一些比较强大的牌型，正在尽力辅助他的队友成为头游，因此他可能会打出手中任何强大的牌来压制我们，然后用单牌辅助他队友获胜。

\medskip

玩家索引为 0，3（敌人）当前对我的手牌的潜在看法：
敌人可能认为我手牌中拥有一些对子和单牌，但具体牌型难以确定。他们可能认为我缺少一些比较强大的牌型，但也可能会警惕我手中的逢人配，需要我在出牌时谨慎选择。

\end{CJK}
\end{mybox}%

\subsection{Related Works}

\subsubsection{Related Works for Large language Models}
Recent research has conducted thorough and comprehensive assessments of the effectiveness of instruction following LLMs' on various tasks~\cite{DBLP:journals/corr/abs-2303-08774, openai2022chatgpt, alpaca, DBLP:journals/corr/abs-2305-12870, baichuan2, chatglm, qwen}, demonstrating its exceptional capability in zero-shot scenarios compared to other models~\cite{
DBLP:journals/corr/abs-2303-12712,
DBLP:conf/emnlp/ChengQCFWCRGZSZ23,
DBLP:journals/corr/abs-2401-07286,
jiayang2024eventground,
DBLP:journals/corr/abs-2309-08303}. 
Nonetheless, there exist persistent challenges that remain unattended, such as the incapacity to engage in intricate mathematical reasoning~\cite{DBLP:journals/corr/abs-2301-13867}, text-to-table generation~\cite{DBLP:journals/corr/abs-2404-14215}, implicit discourse relation recognition~\cite{DBLP:conf/acl/ChanLCLSWS23}, address associated ethical concerns, and tackle privacy issues~\cite{DBLP:journals/corr/abs-2310-10383, DBLP:journals/corr/abs-2212-09292,  DBLP:journals/corr/abs-2311-04044, DBLP:journals/corr/abs-2302-00539,DBLP:journals/corr/abs-2405-07667}.
Therefore, it is critical to discuss whether LLMs based agents possess the capacity of the theory of mind to enhance the performance on  
a multi-player cooperative game under an imperfect information environment.

\end{document}